\PassOptionsToPackage{hyphens}{url} 
\documentclass[letterpaper]{article} 
\usepackage{aaai2026}  
\makeatletter
\gdef\copyright@on{}
\makeatother
\usepackage[table]{xcolor}
\usepackage{booktabs}
\usepackage{times}  
\usepackage{helvet}  
\usepackage{courier}  
\usepackage{url}      
\usepackage{graphicx} 
\usepackage{natbib}  
\usepackage{wasysym} 
\usepackage{caption} 
\usepackage{algorithm}
\usepackage{algorithmic}

\usepackage{newfloat}
\usepackage{listings}
\DeclareCaptionStyle{ruled}{labelfont=normalfont,labelsep=colon,strut=off} 
\floatstyle{ruled}
\newfloat{listing}{tb}{lst}{}
\floatname{listing}{Listing}
\title{Less Redundancy: Boosting Practicality of Vision Language Model \\ in Walking Assistants}
\author{
    Chongyang Li\textsuperscript{1,2,*},
    Zhiqiang Yuan\textsuperscript{1,*,\ensuremath{\dagger}},
    Hanbo Bi\textsuperscript{2},
    Zexi Jia\textsuperscript{1},
    Jinchao Zhang\textsuperscript{1,\ensuremath{\ddagger}}
}

\affiliations{
    \textsuperscript{1}Pattern Recognition Center, WeChat AI, Tencent Inc, China \\
    \textsuperscript{2}University of Chinese Academy of Sciences, Beijing, China \\
        yuanzhiqiang19@mails.ucas.ac.cn \\
    {\small\itshape
    \textsuperscript{*}Equal contribution
    \quad$\cdot$\quad
    \textsuperscript{\ensuremath{\dagger}}Tech Lead
    \quad$\cdot$\quad
    \textsuperscript{\ensuremath{\ddagger}}Corresponding author}
}

\usepackage{bibentry}

\begin{document}

\maketitle

\begin{abstract}
Approximately 283 million people worldwide live with visual impairments, motivating increasing research into leveraging Visual Language Models (VLMs) to develop effective walking assistance systems for blind and low vision individuals. However, existing VLMs in walking assistant task often have outputs that contain considerable redundancy and extraneous details, adversely affecting users' ability to accurately assess their surroundings. Moreover, these models typically lack the capability to proactively assess environmental risks and  adaptively trigger reminders based on the appropriate scene, leading to excessive temporal redundancy. To mitigate output and temporal redundancy, we propose WalkVLM-LR, a walking assistance model with less redundancy. To reduce output redundancy, we introduce four human-preference-based custom reward functions within the GRPO-based reasoning framework to optimize the output in terms of conciseness, fluency, keyword density, and accuracy, thereby producing more informative and streamlined outputs. To minimize temporal redundancy, we incorporate an environment awareness discriminator, which shares the visual encoder with the VLMs to reduce redundant computations and enhance discriminative efficiency, to make WalkVLM-LR assess scene risk levels and minimize unnecessary reminders. Experimental results demonstrate that our method achieves state-of-the-art performance across all evaluation metrics compared with other models, particularly in output conciseness and less temporal redundancy. Code is released at \textcolor{blue}{https://walkvlm-lr.github.io}.
\end{abstract}


\section{Introduction}

Approximately 283 million people worldwide are affected by visual impairments, including about 39 million who are completely blind and 228 million with varying degrees of visual disabilities \cite{world2024global,brady2013visual}. Due to the lack of effective outdoor walking assistive methods, 80\% to 90\% of blind and low vision (BLV) persons report that their daily activities mainly occur indoors \cite{ivanov2010indoor,gamage2023blind}. With the rapid advancement of artificial intelligence (AI) technologies, AI-based walking assistance methods are anticipated to make a substantial contribution to mitigating this issue \cite{real2019navigation,zahn2022obstacle}.

\begin{figure}[!t]
    \centering
     \includegraphics[width=1\linewidth]{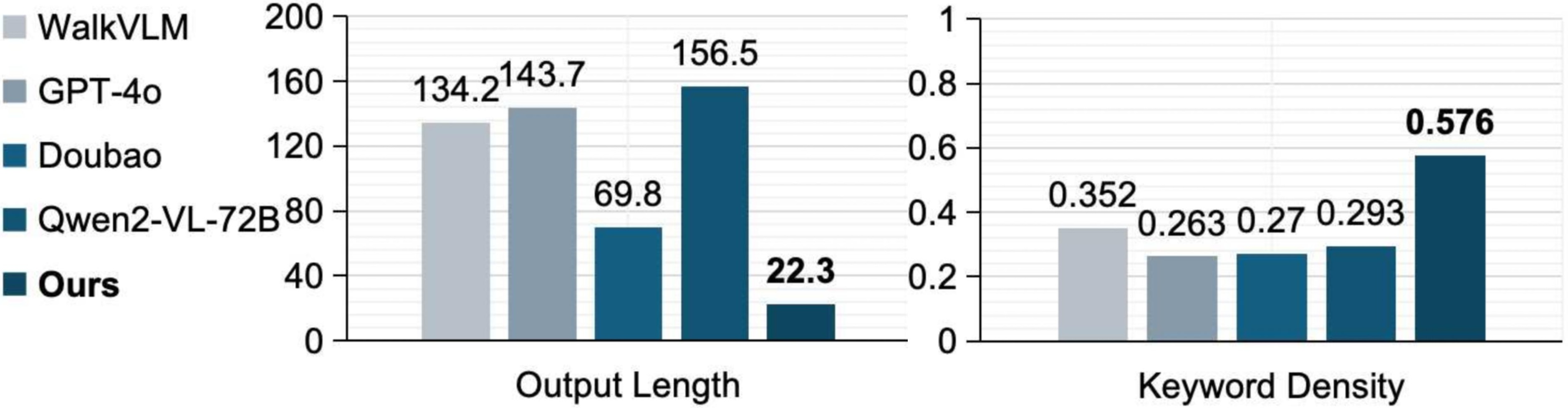}
     \vspace{-18pt}
     \caption{
 The comparison between WalkVLM-LR and mainstream VLMs in terms of output length and keyword density. The outputs of WalkVLM-LR are the shortest and exhibit the highest keyword density, containing the least amount of redundant information.
     }
     \label{fig0}
\vspace{-20pt}
 \end{figure}

Early approaches to walking assistance for BLV individuals primarily relied on visual analysis \cite{liu2023open,okolo2025smart,lee2024multi}. These methods leveraged image understanding techniques to detect and analyze objects in the environment, followed by generating verbal reminders through a language model. However, such approaches are highly dependent on the precision of visual perception and often suffer from low inference efficiency due to their two-stage architecture \cite{abidi2024comprehensive,messaoudi2022review}. In recent years, Vision-Language Model (VLM)-based methods \cite{merchant2024generating,yang2024viassist} have gained significant attention. Yuan $et al.$ \cite{yuan2024walkvlm} proposed WalkVLM along with a high-quality walking assistance dataset. WalkVLM integrates a chain-of-thought reasoning framework to process image inputs, guiding the model hierarchically to produce high-quality outputs and delivers timely reminders through TAP module. Hao $et al.$ \cite{hao2024multimodal} combined image recognition with prompt engineering to generate detailed environmental descriptions and identify potential hazards using VLMs, thereby offering walking assistance for the BLV community.


Although recent progress in VLM-based walking assistance methods has been encouraging, two major challenges persist: output redundancy and temporal redundancy. As shown in the statistical data of Figure \ref{fig0}, the output length of existing VLMs is excessively long, with a relatively low keyword density, indicating a substantial presence of redundancy in the output. The visualization results in Figure \ref{fig3} further reveal that VLMs tend to offer detailed descriptions of the image, rather than providing concise reminders specific to the walking assistant task. This issue arises because VLMs are typically pretrained on image captioning tasks, which emphasize exhaustive visual descriptions \cite{adler2025redundancy,song2024sleb}. However, during downstream fine-tuning, they fail to acquire human preferences specific to the walking assistance domain, resulting in outputs that do not align with task requirements and contain a large amount of redundant information. Additionally, current walking-assistance VLMs still exhibit temporal redundancy and employ inefficient methods for determining the timing of reminder triggers. While some studies \cite{Chen2024_VideoLLMonline, Qian2025_Dispider} have attempted to enable real-time reminder generation using VLMs, they typically delegate the decision-making process to the VLM itself, introducing significant latency and computational overhead.

To mitigate output and temporal redundancy, we present WalkVLM-LR, a walking-assistance VLM optimized for less redundancy. To alleviate output redundancy and enable VLMs to better capture human preferences relevant to walking assistant task, we design four customized reward functions under the GRPO \cite{shao2024deepseekmath} fine-tuning framework. These reward functions guide the model’s output across four dimensions: conciseness, fluency, keyword density, and accuracy. Unlike traditional supervised fine-tuning based on static data and fixed objectives, GRPO leverages dynamic feedback and multi-dimensional, customizable rewards to better align VLM outputs with human preferences. The results in Figure \ref{fig0} demonstrate that, WalkVLM-LR generates the shortest output while achieving the highest keyword density, indicating that our model exhibits the lowest output redundancy. To reduce temporal redundancy and improve the efficiency of trigger timing judgment, we introduce the Environment Awareness Discriminator (EAD). Embedded between the visual encoder and the language model of the VLM, the EAD assesses the danger level of the current scene based on the visual features from the visual encoder and determines whether a reminder is needed, thereby preventing the generation of unnecessary reminders. By sharing the visual encoder with the VLM, the EAD enables trigger timing decisions to be made solely through its own computation, eliminating the need for the language model to perform reasoning, which reduces redundant calculations and enhances discriminative efficiency.

Our main contributions can be summarized as follows:
\begin{itemize}

    \item We propose WalkVLM-LR, a VLM designed for walking assistance, which jointly reduce output and temporal redundancy to improve the informativeness and conciseness of reminders for BLV users.

    \item To reduce output redundancy, we introduce four reward functions aligned with human preferences within the GRPO reasoning framework, guiding the VLM to generate concise, fluent,  informative, and accurate outputs.
    
    \item 
    To reduce temporal redundancy and improving discriminative efficiency, we design Environment Awareness Discriminator that estimates scene risk and triggers reminders only when necessary, thereby suppressing redundant reminders.

    \item 
    Through both quantitative and qualitative experiments, we demonstrate that WalkVLM-LR generates outputs with less redundancy that meets the user's needs, while also achieving lower temporal redundancy.
    
\end{itemize}

\section{Related Work}

\begin{figure*}[t]
  \centering
   \includegraphics[width=0.98\linewidth]{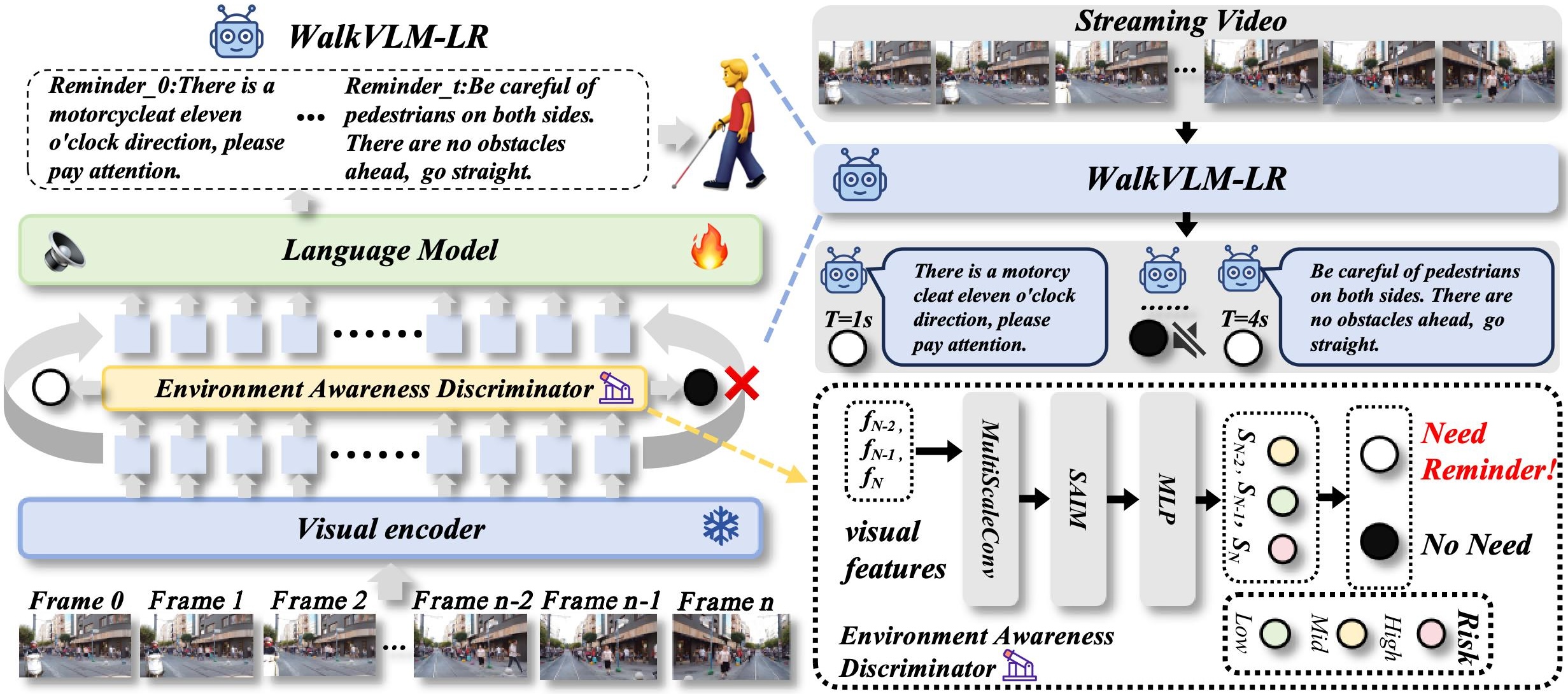}
   \caption{The architecture of WalkVLM-LR. The top-right of the image illustrates how WalkVLM-LR operates, while the left side depicts its overall design. The bottom-right shows the structure of the EAD module. }
   \label{fig1}
   \vspace{-15pt}
\end{figure*} 

\textbf{Existing Methods for Walking Guidance.}
Existing walking guidance methods can be divided into visual analysis based methods and VLM based methods. Visual analysis methods \cite{giudice2020use,gurari2018vizwiz,jain2023want} employ object detection and semantic segmentation to interpret environmental context and deliver guidance through a language model. Liu $et\ al.$ developed OpenSU by integrating GSR \cite{pratt2020grounded} and SAM \cite{kirillov2023segment} to enhance object localization via pixel-level segmentation for BLV individuals. Okolo $et\ al.$ proposed a smart assistive navigation system for BLV individuals that combines object detection with ultrasonic and moisture sensors, providing real-time audio and haptic feedback for enhanced mobility. Lee $et\ al.$ developed a multimodal walking safety system for BLV users, combining YOLOv5-based \cite{khanam2024yolov5} object detection with KoAlpaca-based \cite{yun2023fine} language generation to deliver real-time contextual assistance via voice feedback. Visual analysis methods demand high detection and segmentation accuracy, while their two-stage architecture hinders inference speed, limiting real-time assistive performance.

VLM-based methods integrate visual and linguistic modalities to interpret the environment, using pre-trained VLMs to generate image-grounded textual descriptions. Their end-to-end architecture supports real-time and efficient assistance \cite{chavan2024vocaleyes,huang2025egocentric,baig2024ai}. Merchant $et\ al.$ proposed a system using large language models (LLMs) and VLMs to generate contextually relevant instructions for walking assistants for BLV users. Yang $et\ al.$ proposed VIAssist, a multi-modal LLM system that improves visual question answering for BLV individuals by offering detailed, actionable suggestions for retaking photos and providing reliable answers based on captured images. Yuan $et\ al.$ introduced WalkVLM, a VLM for BLV walking assistance that leverages a large-scale Walking Awareness Dataset and integrates hierarchical reasoning with temporal-aware prediction to provide concise and timely assistance. Although VLM based methods have made significant progress, there is still room for improvement in the timing of generating reminders and the quality of the generated outputs \cite{karamolegkou2025evaluating}.

\noindent
\textbf{Group Relative Policy Optimization.}
Supervised fine-tuning (SFT) has traditionally dominated VLM optimization \cite{jin2024efficient,xing2024survey,zhang2024vision}, but its lack of environmental interaction limits adaptability in dynamic tasks. Recent work explores reinforcement learning to enhance alignment with human preferences and reasoning capabilities. Notably, DeepSeek-AI introduced GRPO in training DeepSeek-R1, integrating chain-of-thought reasoning \cite{wei2022chain} to improve decision-making and achieve strong performance on diverse downstream tasks. Yang $et\ al.$ \cite{yang2024qwen2} applied GRPO to fine-tune Qwen2.5-Math-7B \cite{yang2024qwen25mathtechnicalreportmathematical}, achieving substantial gains on multiple mathematical reasoning benchmarks and demonstrating GRPO's effectiveness in this domain. Alan $et al.$ \cite{dao2025alphamaze} proposed AlphaMaze, leveraging GRPO to improve spatial reasoning in LLMs, and achieved 93\% accuracy on synthetic maze tasks, surpassing traditional approaches. Xue $et al.$ \cite{xue2025dancegrpo} introduced DanceGRPO, applying GRPO to visual generation tasks and achieving improved quality and diversity across multiple benchmarks. The GRPO fine-tuning approach has demonstrated significant and consistent improvements in the reasoning abilities and decision-making quality of VLMs across various domains.

\section{Methods}
The overall structure of WalkVLM-LR is shown in Figure \ref{fig1}. We first introduce the problem formulation, then discuss the four reward functions customized within the GRPO framework, which guide the output to ensure substantive and conciseness while minimizing redundancy. Finally, we present the proposed Environment Awareness Discriminator (EAD), designed to automatically determine the optimal timing for reminder triggers, thereby reducing temporal redundancy.

\begin{figure*}[t]
  \centering
   \includegraphics[width=1.0\linewidth]{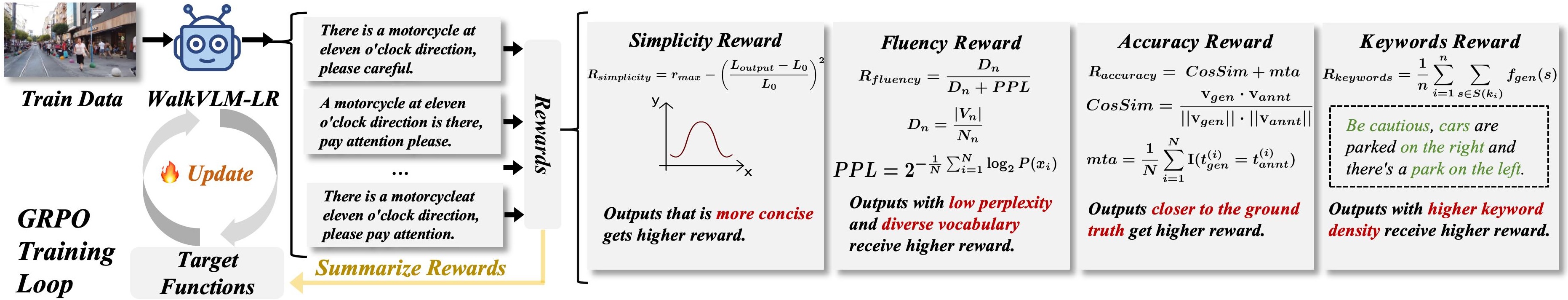}
   \caption{Four Rewards in GRPO Training. GRPO utilizes an intra-group relative advantage evaluation mechanism, combined with four customized reward functions, to assist the VLM in learning how to generate outputs that align with human preferences. }
   \label{fig2}
   \vspace{-15pt}
\end{figure*} 

\subsection{Problem Formulation}
We aim to develop a VLM capable of processing video streams, which can autonomously assess the danger level of the current scene and provide informative and streamlined reminders when necessary. Specifically, at time \( t_n \), the VLM evaluates the danger level of the current scene based on the visual information from the current frame and the preceding \( N \) frames $[ f_{t_{n-N}} ..., f_{t_{n-1}}, f_{t_{n}} ] $. If the danger level of the scene is deemed high and a reminder is required, the model triggers the language model to generate the appropriate reminder $R_{t_n}$. Conversely, if the danger level is low, no reminder is generated. This automatic mechanism for determining when to issue reminders helps to minimize the negative impact of redundant reminders on user experience, while also significantly reducing the real-time processing burden on hardware devices. In addition, to mitigate output redundancy, we design four human-aligned reward functions within the GRPO reasoning framework, aiming to enhance output conciseness, fluency, keyword density, and accuracy.

\vspace{-5pt}
\subsection{Reducing Output Redundancy with GRPO}

To reduce output redundancy while ensuring fluency and accuracy, we fine-tune WalkVLM-LR using GRPO. Traditional SFT merely minimizes the explicit distance between the VLM’s output and reference annotations, which fails to capture human preferences effectively. As illustrated in the Figure \ref{fig2}, GRPO replaces conventional value functions with an intra-group relative ranking mechanism and introduces a flexible multi-objective reward framework, enabling VLMs to align more closely with human preferences in output generation. We design four tailored reward functions to guide the model’s outputs in terms of accuracy, fluency, conciseness, and keyword density. The framework penalizes outputs that are overly redundant or lack fluency, while rewarding those that are accurate and rich in relevant keywords. The following section details the design rationale behind each of the four reward functions.

\subsubsection{Simplicity Reward.}
Simplicity Reward focuses on ensuring that the output is concise, clear, and easy to comprehend, minimizing redundant and repetitive expressions. Excessively long reminders often contain unnecessary redundant information, which hinders visually impaired individuals from quickly understanding the reminders and making timely adjustments. On the other hand, overly short reminders may lead to a loss of essential information, causing key elements to be overlooked. Therefore, we aim for the output length to reach an optimal and balanced level. The closer the generated output length is to this ideal, the greater the reward, while the further it deviates, the smaller the reward. The formula for the Simplicity Reward is as follows:
\begin{equation}
\label{eq:simplicity_reward}
R_{Simplicity} = r_{max} - \left( \frac{{L_{output} - L_0}}{{L_0}} \right)^2
\end{equation}
where $L_{output}$ represents the actual length of the generated output, $L_0$ is the ideal output length, and $r_{max}$ is the maximum reward value.

\subsubsection{Fluency Reward.}
Fluency Reward aims to enhance the grammatical structure and linguistic fluency of the generated output, ensuring that the output adheres to natural language conventions. It consists of two components. The first component measures the perplexity \cite{jelinek1977perplexity} of the output: higher perplexity results in a lower score, indicating poorer fluency. The formula for perplexity is as follows:
\begin{equation}
Perplexity = 2^{-\frac{1}{N} \sum_{i=1}^{N} \log_2 P(x_i)}
\end{equation}
where $N$ represents the total number of words in the output, and $P(x_i)$ is the model's predicted probability for the $i$-th word. The second component evaluates the n-gram \cite{brown1992class} diversity of the generated sentences. Higher n-gram diversity suggests a lower degree of redundancy and repetition in the output, contributing to smoother and more natural sentence structures. The formula for n-gram diversity $D_n$ is as follows:
\begin{equation}
D_n = \frac{|V_n|}{N_n}
\end{equation}
where $|V_n|$ is the number of distinct n-grams in the output, and $N_n$ is the total number of n-grams in the output. The final calculation formula for Fluency Reward is as follows:
\begin{equation}
R_{Fluency} = \frac{D_n}{D_n + PPL}
\end{equation}
where $PPL$ stands for perplexity.

\subsubsection{Accuracy Reward.}
Accuracy Reward is specifically designed to ensure the consistency between the generated output and the expected result. It evaluates the similarity between the generated sentence and the annotation from two key perspectives: semantic similarity and strict matching. Semantic similarity is obtained by encoding both the generated output and the annotation using a pre-trained text encoder and then calculating the cosine similarity. The calculation formula is as follows:
\begin{equation}
{CosSim} = \frac{\mathbf{v}_{gen} \cdot \mathbf{v}_{annt}}{||\mathbf{v}_{gen}|| \cdot ||\mathbf{v}_{annt}||}
\end{equation}
where $\mathbf{v}_{gen}$ is the vector representation of the generated output, and $\mathbf{v}_{annt}$ is the vector representation of the annotation, while $\left\| \bullet \right\|
$ are the magnitudes of these respective vectors.
Strict matching, on the other hand, is evaluated by calculating the mean token accuracy (mta), which involves word-by-word matching between the generated outputs and annotations and calculating the overall accuracy. The calculation formula is as follows:
\begin{equation}
{mta} = \frac{1}{N} \sum_{i=1}^{N} \mathbf{I}(t_{gen}^{(i)} = t_{annt}^{(i)})
\end{equation}
where $N$ is the total number of tokens in the output, $t_{gen}^{(i)}$ is the $i$-th token in the generated output, and $t_{annt}^{(i)}$ is the $i$-th token in the annotation, while $\mathbf{I}$ is an indicator function that returns 1 if the tokens match, and 0 otherwise.
The final Accuracy Reward is the sum of these two values:
\begin{equation}
R_{Accuracy} = CosSim + {mta}
\end{equation}

\subsubsection{Keywords Reward.}
Keywords Reward emphasizes the presentation of key information, ensuring that the output covers the core elements of the task. Keywords Reward is calculated by determining the frequency with which synonyms of the keywords in the annotation appear in the generated output. Specifically, we first extract the keywords from the annotation and identify their synonyms. Then, we calculate the frequency with which these synonyms appear in the generated output. The higher the frequency of these synonyms, the greater the Keywords Reward; conversely, the lower the frequency, the smaller the Keywords Reward. Let $K = \{k_1, k_2, \dots, k_n\}$ be the set of extracted keywords from the annotation, and let $S(k_i)$ denote the set of synonyms for keyword $k_i$. The calculation formula is as follows:
\begin{equation}
R_{Keywords} = \frac{1}{n}\sum_{i=1}^{n} \sum_{s \in S(k_i)} f_{gen}(s)
\end{equation}
where $f_{gen}(s)$ is the frequency of synonym $s$ appearing in the generated output.

These four reward functions work together to ensure that WalkVLM-LR reduces output redundancy while maintaining the fluency and accuracy of the output, all while preserving a high keyword density.

\subsection{Reducing Temporal Redundancy with EAD}
Existing VLMs for walking assistance still exhibit certain temporal redundancy, and the methods for determining the timing for triggering reminders are relatively inefficient. Some studies utilize VLMs to decide when to trigger reminders; however, the full inference process of VLMs tends to be time-consuming, introducing substantial delays. Other studies employ smaller models to assess the danger levels of scenes, but this approach leads to increased computational overhead due to the repeated encoding of visual information. 

To address this issue, we propose the Environment Awareness Discriminator (EAD). As shown in Figure \ref{fig1}, EAD is embedded between the WalkVLM-LR visual encoder and the large language model, automatically determining the danger level of the current scene based on the visual features extracted by the visual encoder. When the scene's danger level is high, the large language model is triggered to generate walking reminders; conversely, if the danger level is low, the language model is not invoked, thus avoiding unnecessary output and reducing temporal redundancy. EAD shares the visual encoder with the VLM, thereby avoiding redundant encoding of visual information. At the same time, it allows for direct assessment of the danger level of a scene without the need to invoke the LLM for inference, thus reducing time consumption.

Specifically, EAD first processes the multi-frame visual features $ [ f_{t_{-n}}, ..., f_{t_{-1}}, f_{t} ]$ extracted by the visual encoder, capturing target information at different scales using a multi-scale convolutional network. Then, it employs the Scene Awareness Inference Module (SAIM), which is composed of several layers of stacked transformers \cite{vaswani2017attention}, to perform in-depth reasoning and judgment on these features. Finally, a multi-layer perceptron (MLP) is used to generate the danger level for each frame of the scene $ [ O_{t_{-n}}, ..., O_{t_{-1}}, O_{t} ]$. Based on the danger levels of the current frame and the previous $n$ frames, the system can determine whether the large language model should be invoked to generate a reminder.

To facilitate the assessment of scene risk, EAD defines three distinct danger levels to comprehensively evaluate the risk of the current environment. Subsequent experiments demonstrate that EAD effectively determines when to trigger reminders, thereby reducing unnecessary outputs. Compared to other VLMs, WalkVLM-LR exhibits a significant reduction in temporal redundancy.

\section{Experiments}
\subsection{Settings}

\begin{table*}[ht]
\centering
\begin{tabular}{l|c|ccccccc}
\hline
\textbf{Model} & Parameters&   ROUGE-1 & ROUGE-2 & ROUGE-L & Keyword Density & GPT Score  \\ \hline
\rowcolor{gray!12} *Qwen2.5-VL-7B-Instruct  &7B& 0.102 & 0.026 & 0.075 & 0.243 & 0.375 \\ 
\/*MiMo-VL-7B-SFT  &7B& 0.086 & 0.026 & 0.069 & 0.251 & 0.433 \\
\rowcolor{gray!12} *InternVL3-8B-Instruct  &8B& 0.118 & 0.033 & 0.094 & 0.293 & 0.417 \\ 
\/*MiniCPM-V-2-6  &8B& 0.091 & 0.026 & 0.072 & 0.261 & 0.351 \\
\rowcolor{gray!12} *Qwen2-VL-72B-Instruct  &72B& 0.107 & 0.028 & 0.079 & 0.293 & 0.363 \\ 
\/*Doubao-seed-1-6-flash  &-& 0.140 & 0.029 & 0.105 & 0.270 & 0.537 \\
\rowcolor{gray!12} *GPT-4o  &-& 0.145 & 0.032 & 0.087 & 0.263 & - \\ 
\/*Gemini-2.5-Flash  &-& 0.117 & 0.024 & 0.079 & 0.216 & 0.489  \\ \hline
\rowcolor{gray!12} Qwen2-VL-2B-Instruct  &2B &0.422 & 0.279 & 0.390 & 0.447 & 0.643 \\ 
Qwen2.5-VL-3B-Instruct  &3B& 0.420 & 0.277 & 0.383 & 0.452 & 0.671  \\ 
\rowcolor{gray!12}Ovis2-2B  &2B& 0.429 & 0.273 & 0.391 & 0.448 & 0.742  \\ 
Deepseek-VL2-Small &2.8B& \underline{0.431} & 0.285 & 0.390 & 0.447 & \underline{0.755}  \\ 
\rowcolor{gray!12}Kimi-VL-A3B-Instruct  &2.8B& 0.428 & 0.289 & \underline{0.393} & \underline{0.459} & 0.662  \\ 
mPLUG-Owl3-2B &2B& 0.401 & 0.264 & 0.366 & 0.445 & 0.712  \\ 
\rowcolor{gray!12}InternVL3-2B  &2B& 0.430 & \underline{0.290} & 0.391 & 0.453 & 0.644  \\ 
InternVL3-1B-Instruct  &1B& 0.417 & 0.280 & 0.381 & 0.439 & 0.636  \\ \hline
\rowcolor{gray!12} WalkVLM-LR &2B& \textbf{0.445} & \textbf{0.310} & \textbf{0.406} & \textbf{0.576} & \textbf{0.776}  \\ \hline
\end{tabular}
\vspace{-5pt}
\caption{Quantitative comparison of different methods. WalkVLM-LR achieves the best performance across ROUGE, Keyword Density, and GPT Score metrics. * denotes zero-shot inference. \textbf{Bold} and \underline{underlined} indicate the best and second-best results.}
\label{Quantitative}
\vspace{-5pt}
\end{table*}

\begin{table*}[!t]
\centering
\setlength{\tabcolsep}{4pt}
\begin{tabular}{l|>{\columncolor{gray!12}}cc>{\columncolor{gray!12}}cc>{\columncolor{gray!12}}cc>{\columncolor{gray!12}}cc>{\columncolor{gray!12}}c}
\hline
\textbf{Model}  & Ovis2  & Owl3 &DeepSeek-VL2 & Qwen2-VL   & InternVL3 & GPT-4o  & Doubao & WalkVLM  & WalkVLM-LR \\ \hline
TRF    & 0.319  & 0.352 &0.427 & 0.449     & 0.475 & 0.430  & 0.459  & \underline{0.505} & \textbf{0.670}
\\ \hline
\end{tabular}
\vspace{-5pt}
\caption{
Evaluation of the temporal redundancy. WalkVLM-LR achieved the highest TRF score, demonstrating its superiority in low temporal redundancy. \textbf{Bold} and \underline{underlined} indicate the best and second-best results.
}
\label{trf}
\vspace{-10pt}
\end{table*}

\begin{table}[!t]
\centering
\setlength{\tabcolsep}{3pt}
\fontsize{9.5pt}{11.5pt}\selectfont
\begin{tabular}{l|ccc}
\hline
\textbf{Model}   &  ROUGE-L & Keyword Density & GPT Score \\ \hline
\rowcolor{gray!12} DeepSeek-VL2          & 0.162 & \underline{0.268} & 0.443   \\
Qwen2-VL    &  0.154  & 0.253 & \underline{0.469}   \\
\rowcolor{gray!12} InternVL3 &  \underline{0.171} & 0.259 & 0.390   \\ 
GPT-4o &  0.063  & 0.226 & -   \\  \hline
WalkVLM-LR  & \textbf{0.214}  & \textbf{0.312} & \textbf{0.568}   \\ \hline
\end{tabular}
\vspace{-8pt}
\caption{
Quantitative comparison on hard samples. WalkVLM-LR achieves the best performance on hard samples across several metrics.
}
\label{hard1}
\vspace{-15pt}
\end{table}

\noindent
\textbf{Models \& Details.}
The implementation of WalkVLM-LR is based on Qwen2-VL-2B-Instruct \cite{wang2024qwen2}, a VLM comprising 2 billion parameters. For both training and evaluation, we adopt the WAD dataset, a high-quality and authoritative dataset specifically designed for walking assistance. It provides diverse, annotated real-world data that enables comprehensive scene understanding and decision-making for BLV individuals. A single epoch of GRPO training is performed using the full dataset. For the EAD, the number of historical frames considered when determining the danger level of the current scene is set to 3. We compare WalkVLM-LR against several models of comparable scale, including Qwen2.5-VL-3B-Instruct \cite{bai2025qwen2}, Ovis2 \cite{brandt2008ovis}, DeepSeek-VL2-Small, Owl3 \cite{ye2024mplug}, InternVL3 \cite{zhu2025internvl3}, and Kimi-VL \cite{team2025kimi}. These baseline models are fine-tuned via one epoch of supervised learning prior to evaluation. Furthermore, we conduct zero-shot experiments on several large-scale models, such as Qwen2-VL-72B-Instruct, GPT-4o \cite{hurst2024gpt}, Doubao-seed-1-6-flash \cite{guo2025seed1}, and Gemini-2.5-Flash \cite{gemini2025pushing}.

\noindent
\textbf{Metrics.}
We adopt four metrics to evaluate model performance: (a) ROUGE \cite{lin2004rouge}, which measures word or phrase overlap between generated outputs and reference texts; (b) Keyword Density, which extracts keywords and CLIP-identified synonyms from the reference, then computes their frequency in the output; (c) GPT Score \cite{zheng2023judging}, calculated as the ratio of GPT-4o evaluations on fluency and conciseness between generated outputs and reference texts; and (d) Temporal Redundancy F1 (TRF), which compares predicted and actual danger levels across frames to assess temporal redundancy.

\vspace{-5px}
\subsection{Quantitative Results}
\begin{figure*}[htbp]
  \centering
   \includegraphics[width=1.0\linewidth]{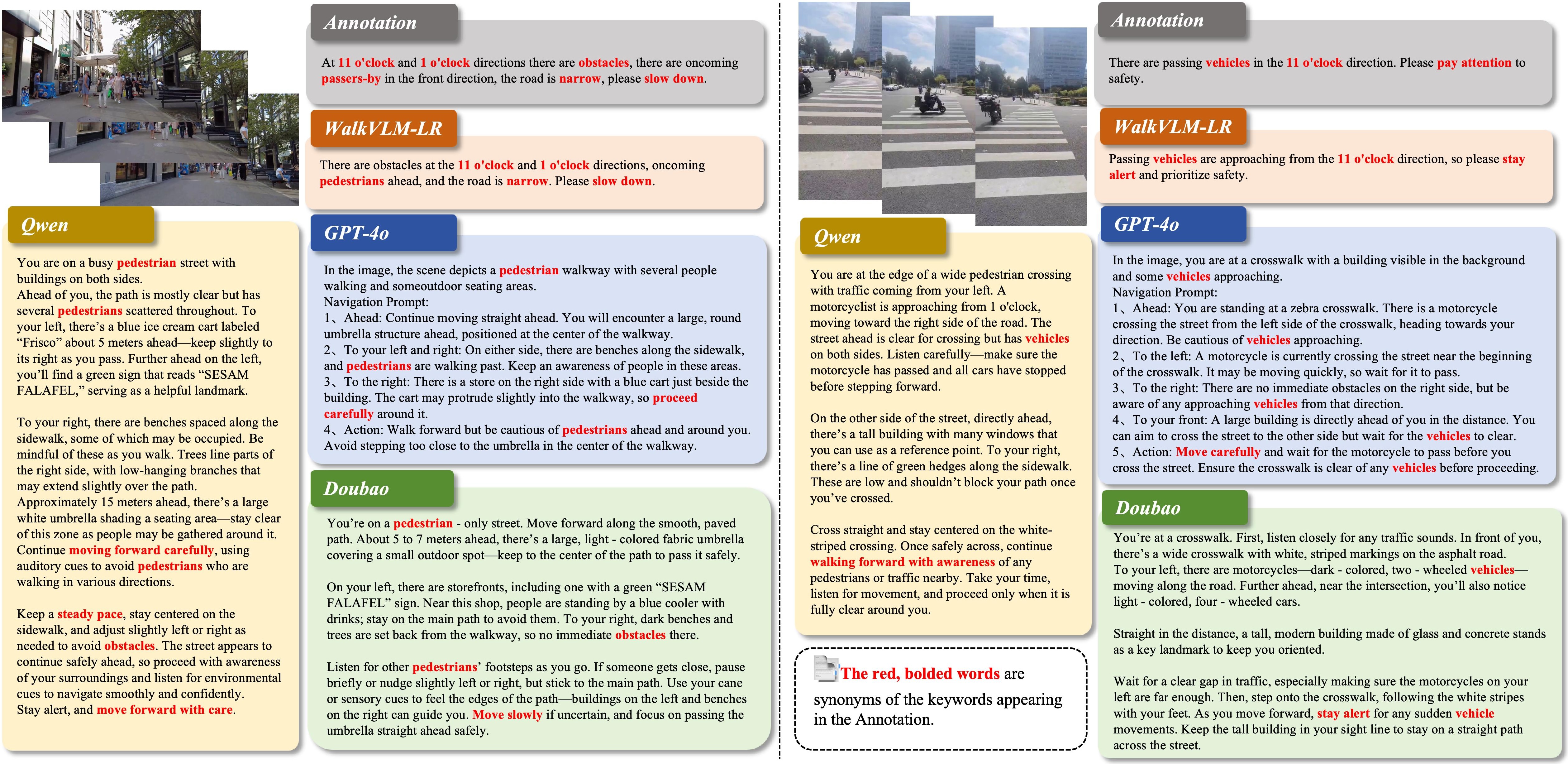}
   \vspace{-15px}
   \caption{Output comparison of different VLM models in walking assistant task. Compared to other models, WalkVLM-LR generates concise and informative responses, offering better guidance for visually impaired users. }
   \label{fig3}
\vspace{-5pt}
\end{figure*} 
\subsubsection{Outputs Comparison.}Table \ref{Quantitative} presents a comprehensive quantitative comparison of different VLMs. The proposed WalkVLM-LR consistently outperforms others across multiple evaluation metrics, including ROUGE, Keyword Density, and GPT Score. In terms of ROUGE, which evaluates the similarity between the generated outputs and reference text, it achieves the best results in ROUGE-1, ROUGE-2, and ROUGE-L, indicating that its outputs are closest to the ground truth. Additionally, the model excels in Keyword Density, demonstrating a high level of lexical similarity with the reference text while effectively conveying key information. The GPT score, which evaluates fluency and conciseness, further underscores its superiority, with a score of 0.776 that significantly exceeds those of larger VLMs with more parameters, highlighting the simplicity and fluency of the generated output. Figure \ref{fig0} compares the output length and keyword density generated by the WalkVLM-LR, WalkVLM, and other mainstream VLMs on the test dataset. As shown, our model generates the most concise output with the highest keyword density. These experimental results clearly demonstrate that WalkVLM-LR effectively reduces output redundancy, producing outputs that meet the specific requirements of the walking assistant task.

\subsubsection{Temporal Comparison.}In addition to the comprehensive quantitative evaluation of the output quality generated by WalkVLM-LR, we also conduct a detailed quantitative assessment of the temporal redundancy of different VLMs. We provide the VLM with both the current scene frame and several previous frames to assess the danger level of the current scene and determine whether a reminder is needed. As shown in Table \ref{trf}, WalkVLM-LR achieves the highest TRF score among all VLMs, underscoring its exceptional ability to effectively reduce temporal redundancy.

\subsubsection{Outputs Comparison on Challenging Scenarios.}
To evaluate WalkVLM-LR's performance in challenging scenarios, we selected a subset of difficult samples from the test set, including complex environments, nighttime scenes, and scenarios with significant changes (selection criteria in the Appendix.\textcolor{red}{B}). A comparison with top-performing VLMs from Table \ref{Quantitative} is presented in Table \ref{hard1}. As shown, all VLMs experienced performance drops across metrics, while WalkVLM-LR consistently achieved the highest scores, demonstrating its robustness in challenging scenarios.

\subsubsection{Temporal Comparison on Challenging Scenarios.}
We further evaluated the temporal redundancy of VLMs on these challenging scenarios. VLMs with better TRF performance in Table \ref{trf} were selected for comparison, and the results are shown in Table \ref{hard2}. WalkVLM-LR achieved the highest TRF score. Despite the inherent difficulty of identifying the correct timing for reminders in complex scenarios, WalkVLM-LR was able to effectively assess the danger level of the current scene and minimize unnecessary reminders.

\vspace{-2pt}
\subsection{Qualitative Results}
Figure \ref{fig3} provides a visual comparison of outputs generated by WalkVLM-LR and other VLMs in the same scenario. Compared to GPT-4o, Qwen, and Doubao, our model produces more concise and fluent outputs with higher keyword density, enabling BLV users to more easily understand the scene and make quicker decisions. In the left example, other VLMs provide detailed but redundant scene descriptions and suggestions, which hinder swift comprehension. In contrast, our model generates concise, key information, improving the user experience. On the right, while other VLMs focus on distant objects, WalkVLM-LR prioritizes immediate surroundings and offers clear warnings about nearby hazards.

\begin{table}[!t]
\centering
\setlength{\tabcolsep}{1pt}
\fontsize{9.5pt}{11.5pt}\selectfont
\begin{tabular}{l>{\columncolor{gray!12}}cc>{\columncolor{gray!12}}cc>{\columncolor{gray!12}}c}
\hline
\textbf{Model}  & Qwen2-VL  & InternVL3 &GPT 4o & Doubao  & WalkVLM-LR \\ \hline
\textbf{TRF}    & 0.394  & \underline{0.450} &0.424 & 0.405 & \textbf{0.663}
\\ \hline
\end{tabular}
\vspace{-8pt}
\caption{
Evaluation of temporal redundancy on hard samples. WalkVLM-LR achieved the highest TRF score.
}
\vspace{-20pt}
\label{hard2}
\end{table}
\vspace{-4pt}
\subsection{Subjective Results}
\begin{figure}[!t]
    \centering
     \includegraphics[width=1\linewidth]{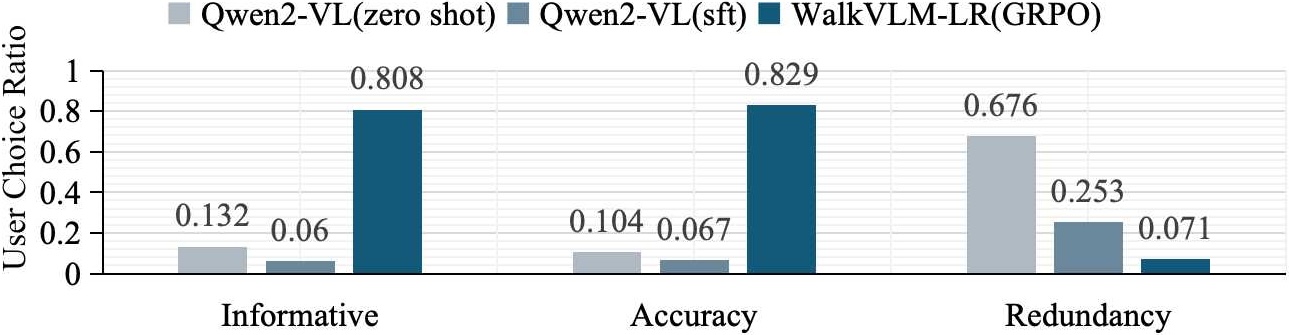}
     \vspace{-20pt}
     \caption{
Results of the user study regarding informativeness, accuracy, and redundancy. Higher scores indicate better performance in terms of informativeness and accuracy, while lower scores are indicative of better performance in terms of redundancy.
     }
     \label{fig6}
     \vspace{-15pt}
 \end{figure}

\vspace{-2pt}
To evaluate the performance of WalkVLM-LR in practical applications, we conducted a user experiment. Seven volunteers were recruited to assess the outputs of different VLMs based on real-world video scenarios, focusing on three aspects: informative, accuracy, and redundancy. For each sample, the volunteers selected the model that performed best in each aspect based on its output. The score of each model for a given metric was calculated by dividing the number of samples in which the model achieved the best performance by the total number of samples. For the informative and accuracy metrics, higher scores indicate better performance, whereas for redundancy, lower scores are preferred. The results, shown in Figure \ref{fig6}, demonstrate that WalkVLM-LR significantly outperforms other models in both informative and accuracy, indicating that its outputs are both accurate and rich in key information. Moreover, WalkVLM-LR exhibits a much lower redundancy compared to the other models. Although Qwen2-VL with supervised fine-tuning shows some improvement in redundancy, the use of a single supervisory signal does not allow it to simultaneously optimize both Informative and Accuracy. On the other hand, the zero-shot trained Qwen2-VL performs well in terms of informative and accuracy, but suffers from excessive redundancy.

\subsection{Ablative Study}
\subsubsection{Rewards Ablation.}Table \ref{aba} summarizes the ablation studies for WalkVLM-LR, evaluating the contribution of each reward function used in GRPO training. Four experiments were conducted by removing one reward at a time while keeping the others fixed. Results show that excluding the Simplicity Reward leads to excessive verbosity and significant declines across all metrics. Removing the Fluency Reward has minimal impact on ROUGE scores but substantially lowers the GPT Score, indicating reduced textual fluency. Without the Accuracy Reward, reliance on only the Keywords Reward yields a sharp drop in accuracy and ROUGE performance. Finally, omitting the Keywords Reward weakens the enforcement of informational relevance, resulting in degraded performance across multiple metrics.

\subsubsection{EAD Ablation.}Table \ref{EAD} presents an extensive ablation study on EAD components. Removing the MultiScaleConv (MSC) module impairs multi-scale object perception, significantly reducing accuracy in reminder timing. Excluding the SAIM module weakens reasoning ability, consequently leading to lower TRF scores. Omitting the MLP module retains basic visual analysis but compromises regression performance, ultimately resulting in less precise predictions.


\begin{table}[!t]
\centering
\setlength{\tabcolsep}{2pt}
\fontsize{9.5pt}{11.5pt}\selectfont
\resizebox{1\linewidth}{!}{
\begin{tabular}{l|cccc}
\hline
Configuration   & ROUGE-1 & ROUGE-2 & ROUGE-L & GPT Score \\ \hline
\rowcolor{gray!12} w/o $R_{Simplicity}$          & 0.370  & 0.229   & 0.345   & 0.469   \\
w/o $R_{Fluency}$    & \underline{0.426}  & \underline{0.266}   & 0.382   & 0.447   \\
\rowcolor{gray!12} w/o $R_{Accuracy}$ & 0.393  & 0.247   & 0.368   & \underline{0.672}   \\ 
w/o $R_{Keywords}$ & 0.418  & 0.254   & \underline{0.393}   & 0.651   \\ \hline
All Rewards             & \textbf{0.445}  & \textbf{0.310}   & \textbf{0.406}   & \textbf{0.776}   \\ \hline
\end{tabular}
}
\vspace{-5pt}
\caption{
Ablation study on GRPO rewards. Each reward function demonstrates its intended effect.
}
\label{aba}
\vspace{-5pt}
\end{table}

\begin{table}[!t]
\centering
\setlength{\tabcolsep}{2pt}
\fontsize{9.5pt}{11.5pt}\selectfont
\begin{tabular}{c|>{\columncolor{gray!12}}cc>{\columncolor{gray!12}}cc}
\hline
Configuration  & w/o MSC  & w/o SAIM & w/o MLP & WalkVLM-LR \\ \hline
TRF    & \underline{0.653}  & 0.637 &0.437 & \textbf{0.670}
\\ \hline
\end{tabular}
\vspace{-8pt}
\caption{
Ablation study on the components of EAD. Each module in EAD contributes to improving TRF.
}
\label{EAD}
\vspace{-15pt}
\end{table}

\section{Conclusion}

This paper introduced WalkVLM-LR, a novel walking assistance VLM specifically designed to address the challenges of output and temporal redundancy in existing VLM-based methods. Through the integration of human-preference-based reward functions and an environment awareness discriminator, WalkVLM-LR demonstrates superior performance in generating concise, fluent, and informative outputs, while also minimizing temporal redundancy. Our comprehensive experimental results show that our method outperforms existing VLMs in terms of output quality and scene risk assessment, making it a promising and effective solution for enhancing walking assistance for BLV individuals.

\section{Acknowledgments}
This work was part of Chongyang Li's research during his internship at Tencent WXG, under the guidance of Zhiqiang Yuan, and both made equivalent contributions.

\bibliography{aaai2026}

\clearpage
\section{Appendix Overview}
The Supplementary Materials provide additional information that complements the main paper, organized into three major sections:
\begin{itemize}
\item \textbf{Training Configurations and Evaluation Settings in Appendix A.}
This section presents the detailed configurations used during the experiments. A.1 describes the supervised fine-tuning (SFT) and GRPO-based reinforcement learning setups, including learning rates, batch sizes, and optimization strategies. A.2 introduces the evaluation metrics used to measure model performance across multiple dimensions. A.3 provides a comprehensive description of the user study setup, outlining participant recruitment, task design, and feedback collection procedures.
\item \textbf{Dataset Description and Sample Selection in Appendix B.}  
This section elaborates on the datasets used in the experiments. B.1 presents a thorough introduction to the data sources and key statistics. B.2 explains the procedure for filtering and selecting challenging samples, highlighting the criteria designed to ensure the evaluation focuses on difficult and representative cases.
\item \textbf{Training Details in Appendix C.}  
This section analyzes the GRPO training dynamics and EAD module training. C.1 examines the evolution of the reward function, KL divergence, and training loss trends, providing insights into model stability. C.2 focuses on the training details of the EAD module.
\item \textbf{Challenges and Future Directions in Appendix D.}  
This section discusses the limitations and prospects of the current work. It analyzes low-performing or failed samples to identify the underlying causes of errors and outlines the challenges in developing an edge-side application, highlighting potential issues in model deployment and the roadmap for a lightweight app implementation.
\end{itemize}

\section{Appendix A: Training Configurations and Evaluation Settings}
\subsection{A1. Training Setups}
\subsubsection{GRPO Training Setups.}
We implement WalkVLM-LR using the OPEN-R1\cite{deeplseek2025r1} framework. The backbone models are based on Qwen2-VL-Instruct \cite{wang2024qwen2} with 2B parameters. The training was conducted on the WAD dataset \cite{yuan2024walkvlm}, with a maximum input sequence length of 1024 tokens and a maximum generation length of 700 tokens. A per-device batch size of 1 was used, with gradient accumulation over 2 steps to effectively increase the training batch size. The model was trained for 1 epochs using bfloat16 mixed-precision \cite{deepmind2020bfloat16} to improve computational efficiency and reduce memory usage. Gradient Checkpointing \cite{chen2021gradient} was enabled to further optimize memory consumption. The attention mechanism was implemented using Flash Attention2 \cite{flash-vl-2b-2025} to accelerate training. The maximum resolution for image inputs was set to 401,408 pixels. Logging was performed at every training step, and model checkpoints were saved every 100 steps, retaining only the model weights. 

\subsubsection{SFT Training Setups.}
We implement comparative experiments using the MS-SWIFT~\cite{zhao2024swiftascalablelightweightinfrastructure} framework. The model was fine-tuned using the LoRA \cite{hu2021lora} method with a rank of 8 and scaling factor of 32, applied to all linear layers while freezing the vision transformer. Training was conducted for 2 epochs with a per-device batch size of 1 and gradient accumulation over 4 steps. The input length and image resolution were limited to 15,536 tokens and 1,003,520 pixels, respectively. Mixed precision training with bfloat16 was used, and DeepSpeed ZeRO \cite{raja2020deepspeed} Stage 2 was applied for memory optimization. The learning rate was set to 1e-4 with a warmup ratio of 0.05. Evaluation and checkpoint saving occurred every 50 steps, keeping up to 2 checkpoints, and only model weights were stored. Data loading and preprocessing used 4 worker threads each.

\subsection{A2. Evaluation Metrics}

We use the following evaluation metrics to comprehensively assess the performance of the model: \textbf{(a) ROUGE}: ROUGE \cite{lin2004rouge} measures the similarity between the generated text and the reference text, primarily by comparing overlapping words or phrases. It includes ROUGE-1, ROUGE-2, and ROUGE-L. (b) \textbf{Keyword Density(KeyDens)}: We first extract keywords from the reference text and then use a pre-trained CLIP \cite{radford2021clip} model to encode synonyms of these keywords. We consider words with a cosine similarity greater than 0.9 as synonyms. The frequency with which these synonyms appear in the generated text is calculated to determine the Keyword Density. \textbf{(c) GPT Score}: GPT-4 \cite{openai2023gpt4} is used to evaluate the fluency and conciseness of both the reference text and the generated text by WalkVLM2.0. The scores from GPT-4 for both the reference and the generated text are then compared, and their ratio is computed to obtain the GPT Score. A higher GPT Score indicates better performance in terms of fluency and conciseness in the generated text.  \textbf{(d) Temporal Redundancy F1 Score (TRF)}: We calculate the F1 score between the predicted danger levels of the historical and current frames and the ground truth to quantify redundancy \cite{chen2022temporal}. A higher TRF value indicates less temporal redundancy in the generated reminders, thus demonstrating greater efficiency. These evaluation metrics cover multiple dimensions of the model's performance: ROUGE assesses the accuracy of the generated text, Keyword Density measures the coverage and proportion of key information in the generated text, and GPT Score evaluates the fluency and conciseness of the generated text by comparing it with the reference text. Additionally, TRF quantifies temporal redundancy, ensuring that the generated reminders are more efficient with lower redundancy.

\subsection{A3. User Study}
\subsubsection{User Study Setups.}
To evaluate model outputs from a human perspective, we conducted a user study based on real-world scenarios. Three videos were collected to represent diverse walking environments: (1) a nighttime pedestrian walkway, (2) a daytime park scene, and (3) a daytime street environment. Each video was sampled at 2-second intervals to generate image frames for evaluation. We compared three models: WalkVLM-LR, the 2B-parameter Qwen2-VL-instruct in zero-shot mode, and the 2B-parameter Qwen2-VL-instruct after supervised fine-tuning (SFT). All models generated outputs for the sampled frames. The outputs were reformatted into multiple-choice questions, where participants evaluated model responses along three criteria: \textbf{informativeness}, \textbf{accuracy}, and \textbf{redundancy}. Seven volunteers were asked to select, for each criterion, the output that best exhibited the corresponding property. We then aggregated the votes across all tasks. A higher vote count in \textbf{informativeness} and \textbf{accuracy} indicates higher output quality, whereas a higher vote count in \textbf{redundancy} suggests lower generation quality. This design allowed us to measure comparative performance across models in a human-centric evaluation.

\section{Appendix B: Dataset Description and Sample Selection}

\subsection{B1. Dataset Description}
The Walking Awareness Dataset (WAD) \cite{yuan2024walkvlm} is a multimodal dataset for blind walking assistance, containing 12k videos / 120k images collected from 10 locations across Europe and Asia. It includes approximately 13 hours of walking footage, with 20\% recorded by annotators and the rest sourced from YouTube. Each sample is annotated with scene attributes (e.g., weather, location type, traffic level, danger rating) and semantic labels, including 6 types of reminders and 3 categories of QA. Annotations are guided by blind test experiments and refined through GPT-assisted rephrasing. Compared to prior datasets, WAD offers greater scale, diversity, and richer annotations, supporting both perception and decision-level tasks in walking assistance.

\subsection{B2. Method for Selecting Hard Examples}
To systematically identify difficult samples for further model analysis and evaluation, we implemented an automatic scene-level difficulty assessment procedure based on a large-scale vision-language model (VLM). Specifically, we designed a danger-level classification task, where each scene frame is evaluated and assigned one of three risk levels: Grade A (low risk), Grade B (medium risk), or Grade C (high risk). The grading criteria consider factors such as visual complexity, dynamic changes, lighting conditions, and potential safety hazards.

We leveraged the Qwen2-VL-72B model, deployed via a custom API, to perform this classification. For each selected frame from the test split of the WAD dataset, an image is input to the model along with a carefully crafted system prompt describing the evaluation criteria. The model's response is restricted to outputting only a single grade (A/B/C) to ensure consistency. This process is applied to all samples using a batch evaluation script, and the predicted danger level is appended to each sample entry for further analysis.

This approach enables the automatic curation of high-risk (Grade C) samples, which are particularly valuable for testing the robustness of walking assistant models under challenging conditions such as low visibility, crowded environments, or scene instability.

\section{Appendix C: Training Details}
\subsection{C1: GRPO Training Dynamics Analysis.}
\begin{figure}[!t]
    \centering
     \includegraphics[width=0.9\linewidth]{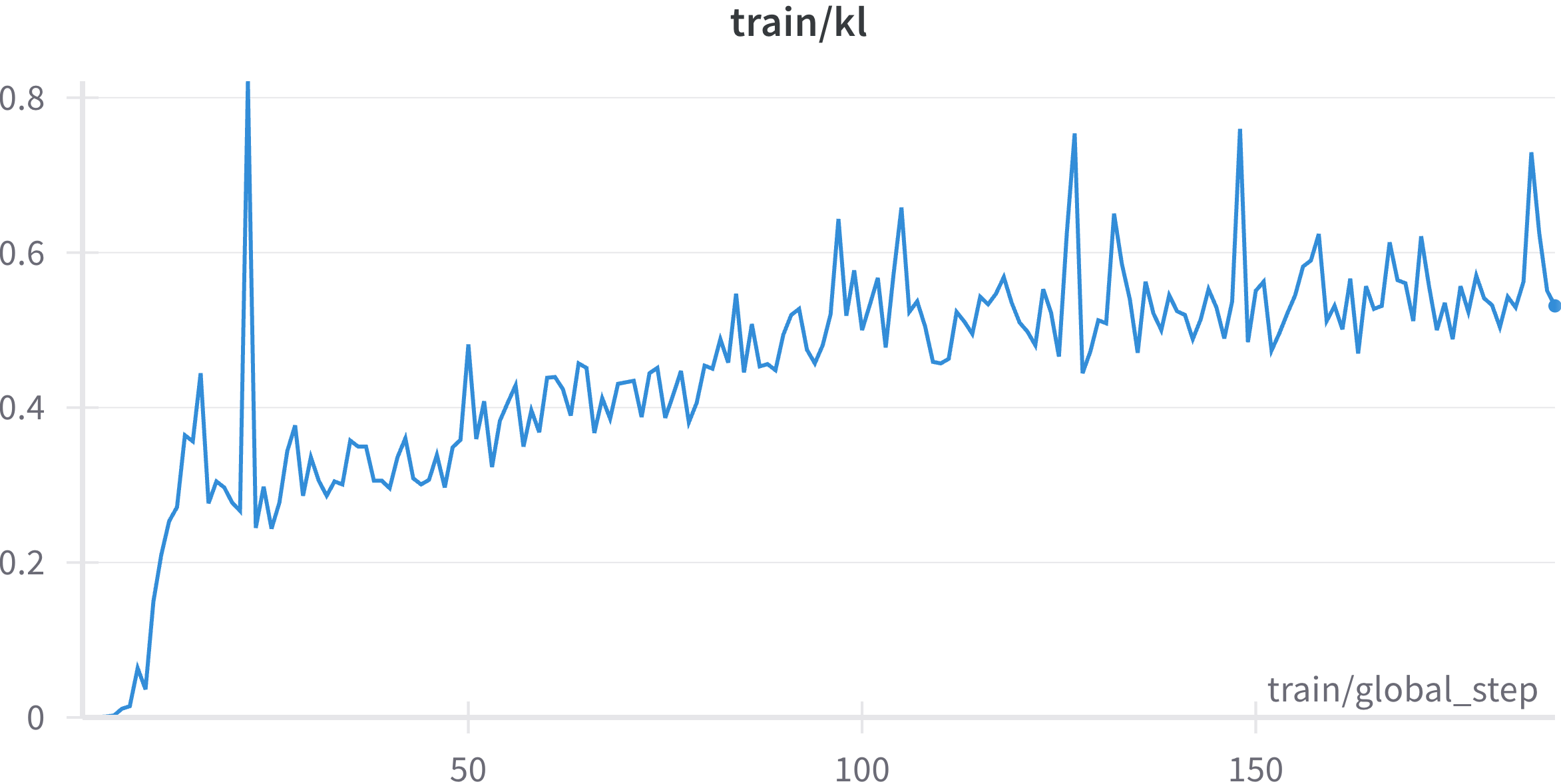}
     \caption{
The variation of KL Divergence during training.
     }
     \label{1}
 \end{figure}

\begin{figure}[!t]
    \centering
     \includegraphics[width=0.9\linewidth]{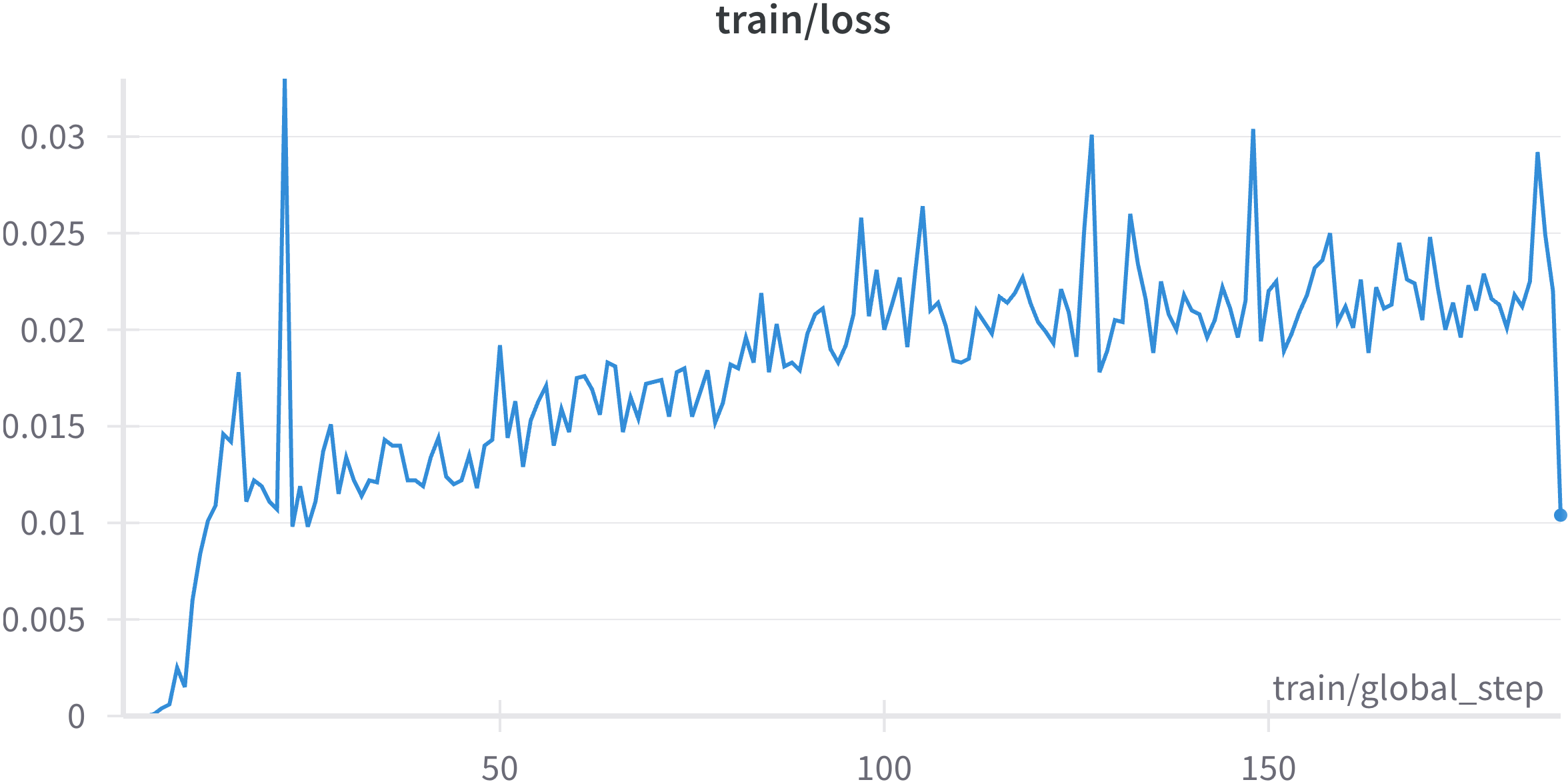}
     \caption{
The variation of Loss during training.
     }
     \label{2}
 \end{figure}

\begin{figure}[!t]
    \centering
     \includegraphics[width=0.9\linewidth]{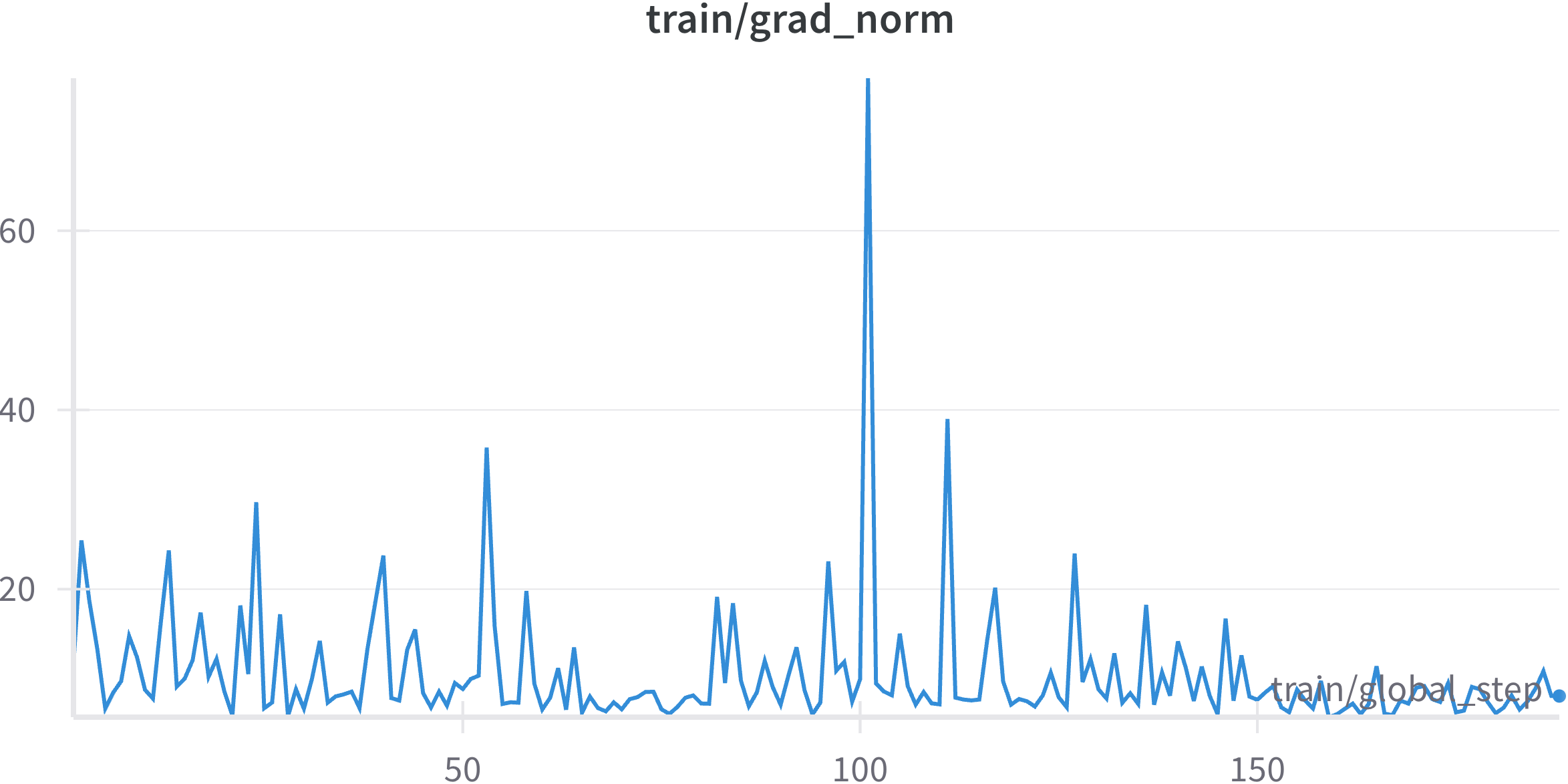}
     \caption{
The variation of Gradient Norm during training.
     }
     \label{3}
 \end{figure}

\begin{figure}[!t]
    \centering
     \includegraphics[width=0.9\linewidth]{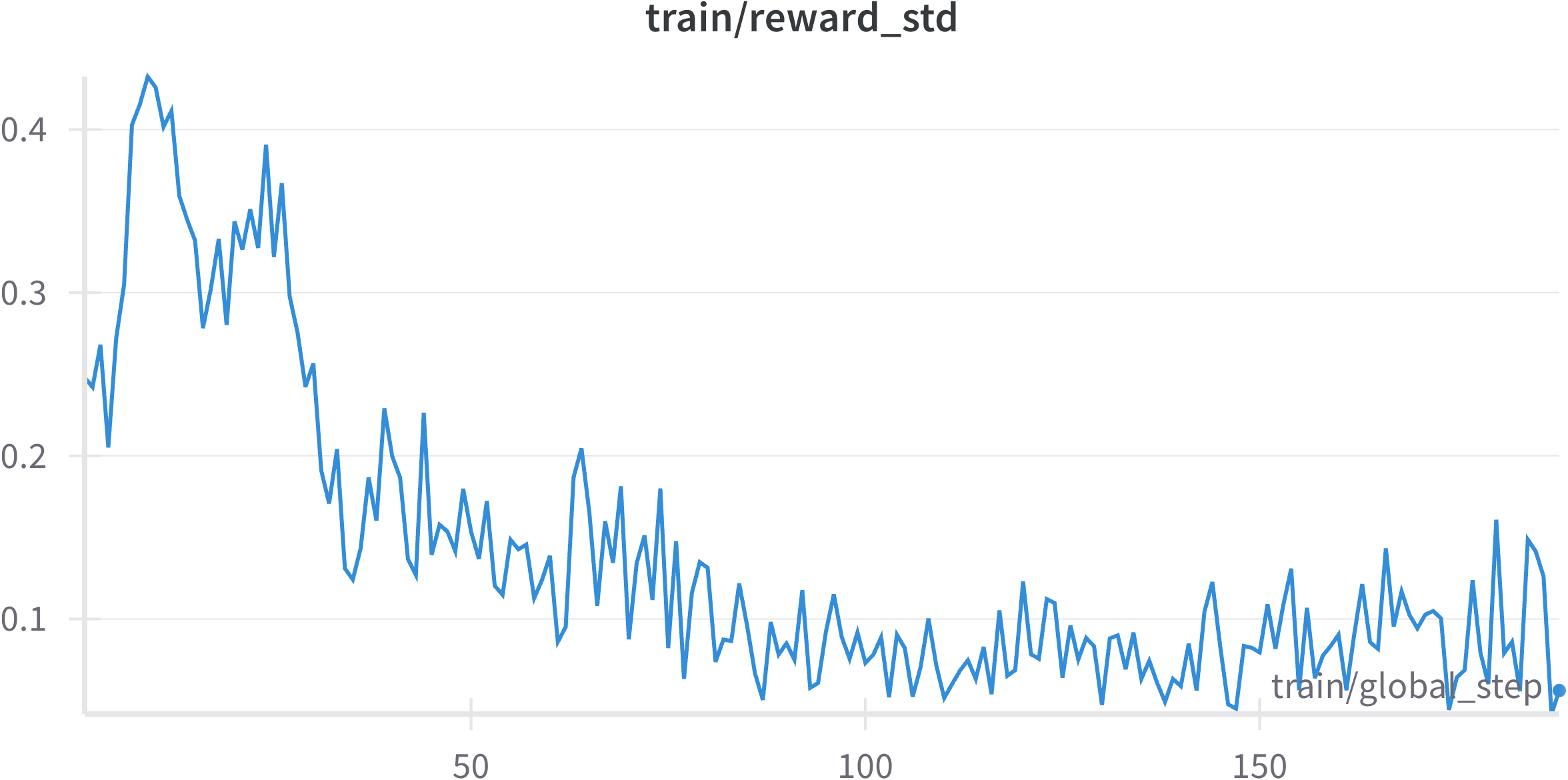}
     \caption{
The variation of Reward Standard Deviation during training.
     }
     \label{4}
 \end{figure}

Figures \ref{1},\ref{2},\ref{3} and \ref{4} illustrate the changes in KL Divergence \cite{kl_divergence}, loss, Gradient Norm \cite{gradient_norm}, and Reward Standard Deviation throughout the training process. From these plots, it can be observed that the KL Divergence exhibits some fluctuations in the early stages of training but gradually stabilizes as the training progresses. This suggests that the model effectively adjusts its policy distribution, progressively aligning with the target distribution, and demonstrates continuous improvement in policy optimization. The rapid decline in the loss curve indicates that the model is able to quickly reduce error in the early stages of training, and further optimizes in subsequent phases, reflecting the model’s strong adaptability and efficient optimization capacity.

Although the Gradient Norm curve shows some fluctuations, it generally trends toward stability, indicating that the model gradually finds a consistent optimization direction and the gradient stability is well controlled, contributing to continued convergence. Finally, the gradual reduction in Reward Standard Deviation is one of the most positive signals in the training process, suggesting that, as training progresses, the model’s decisions become increasingly consistent and stable, with reduced reward variability. This reflects the model’s stability and robustness in policy generation.

Overall, these trend curves demonstrate the stability and efficiency exhibited by the GRPO model throughout the training process. The gradual improvement in various metrics not only indicates that the model is continuously refining and optimizing its policy but also highlights the model’s ongoing progress and excellent performance in tackling complex tasks.


\begin{figure}[htbp]
    \centering
     \includegraphics[width=0.9\linewidth]{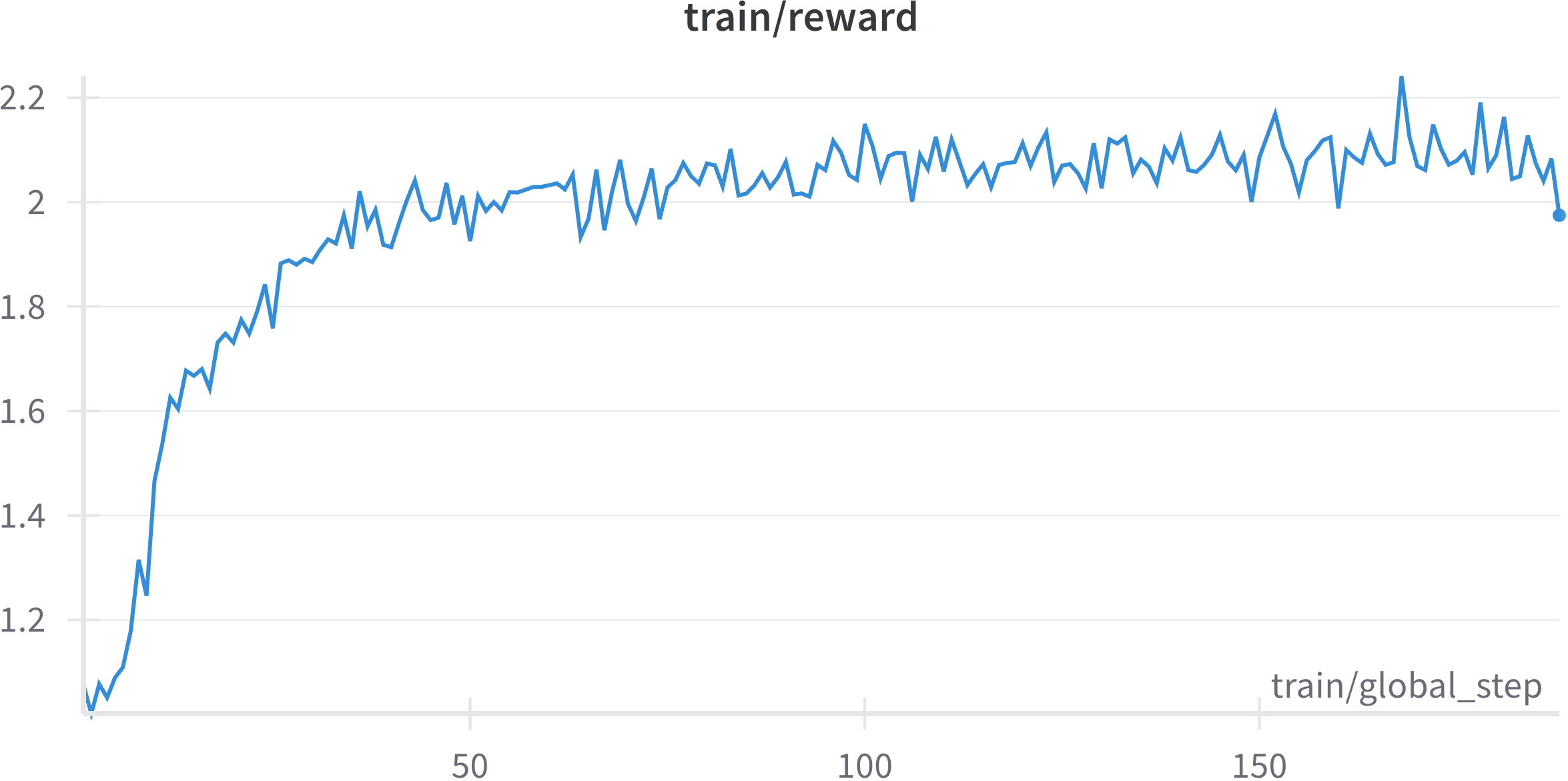}
     \caption{
The variation of Total Reward during training.
     }
     \label{5}
 \end{figure}

\begin{figure}[!t]
    \centering
     \includegraphics[width=0.9\linewidth]{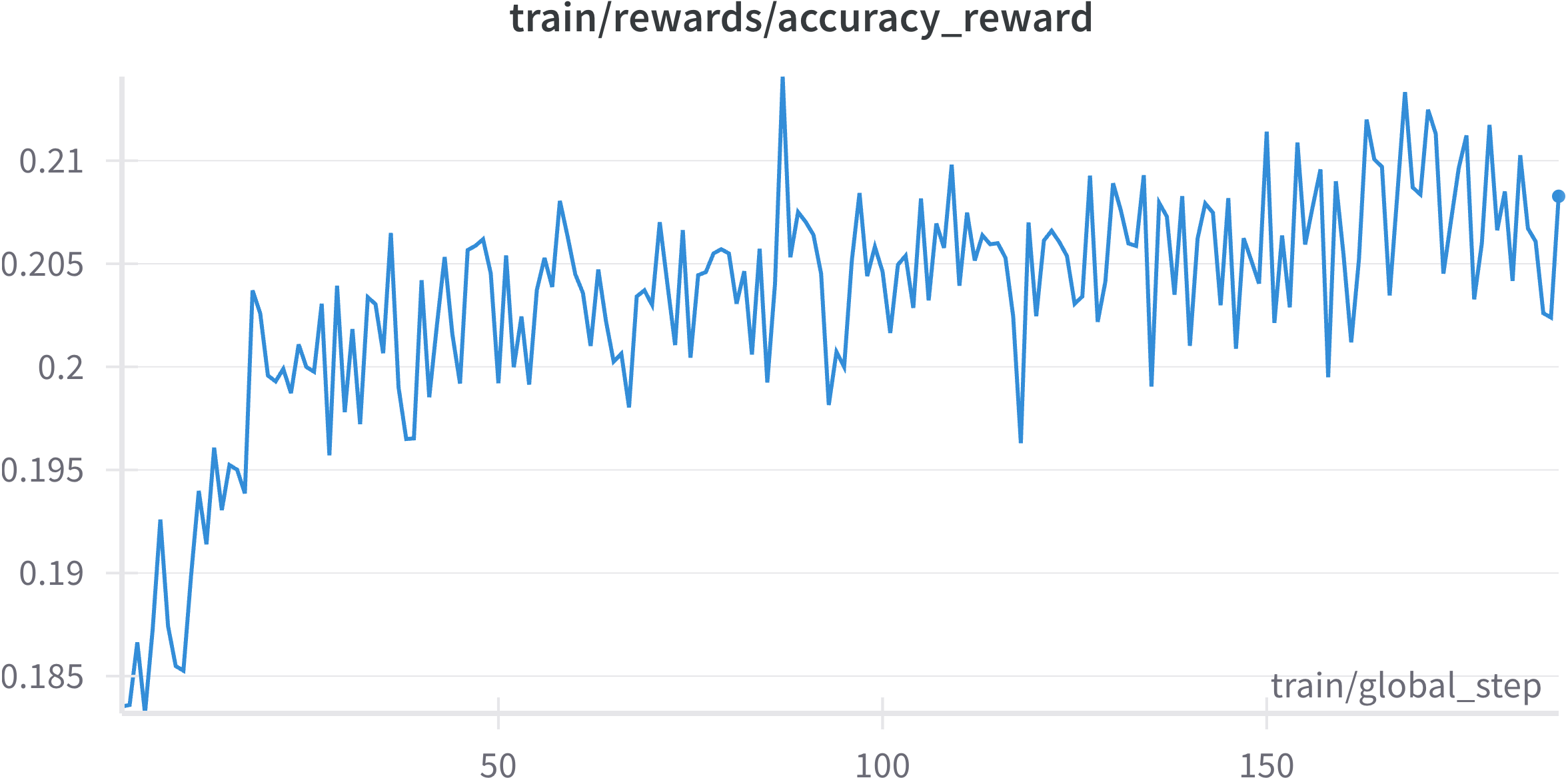}
     \caption{
The variation of Accuracy Reward during training.
     }
     \label{6}
 \end{figure}

\begin{figure}[!t]
    \centering
     \includegraphics[width=0.9\linewidth]{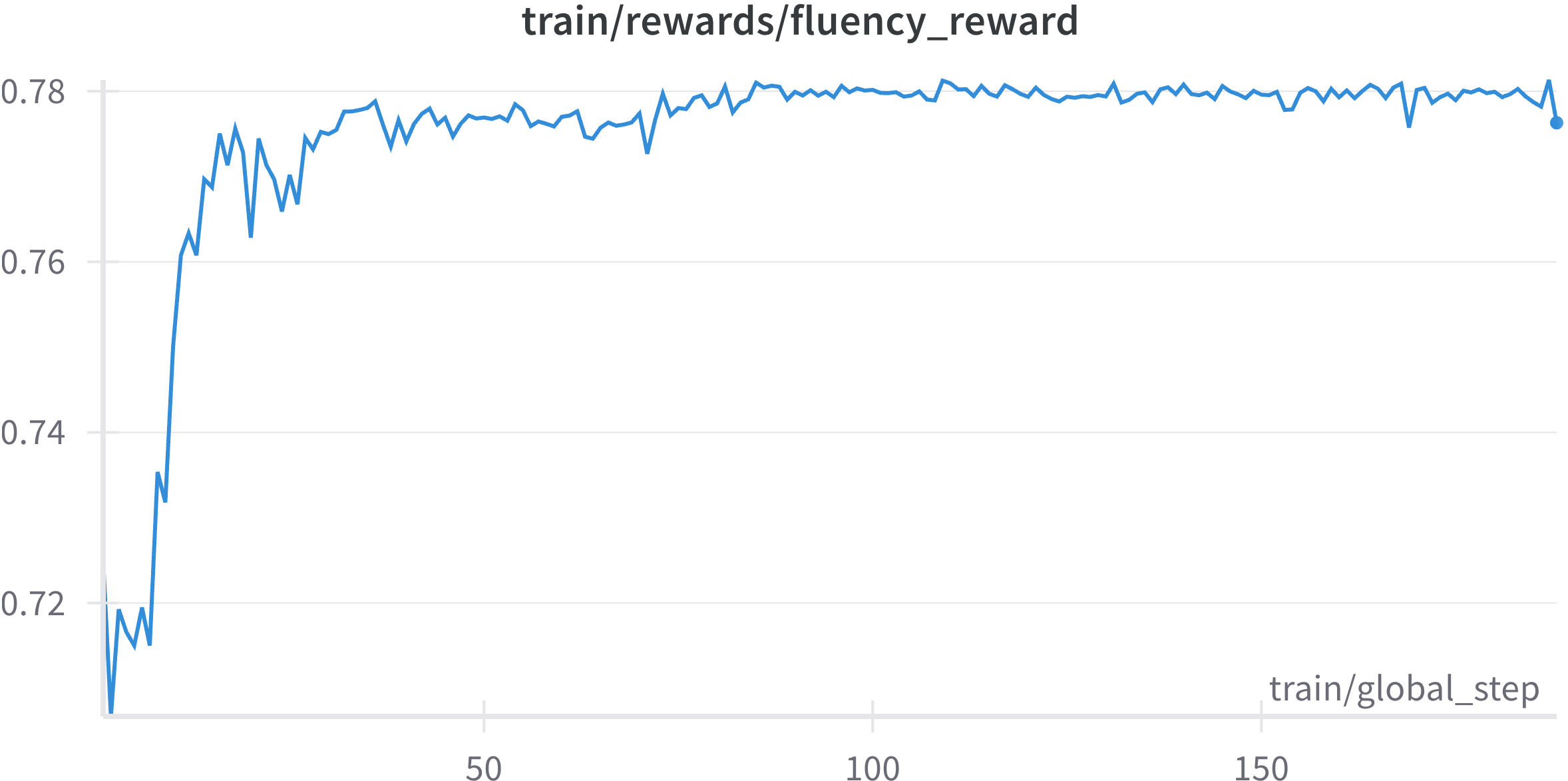}
     \caption{
The variation of Fluency Reward during training.
     }
     \label{7}
 \end{figure}

\begin{figure}[!t]
    \centering
     \includegraphics[width=0.9\linewidth]{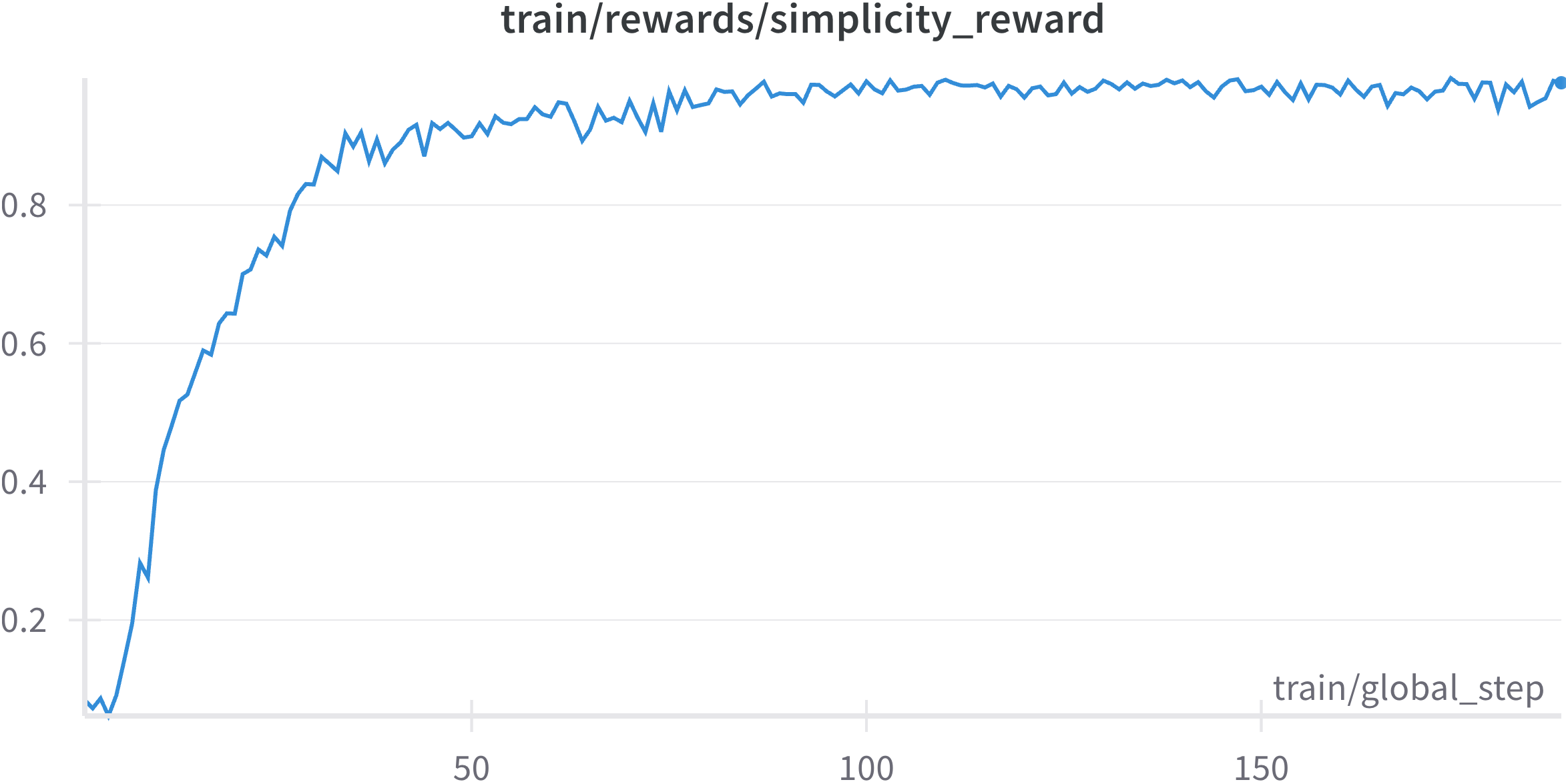}
     \caption{
The variation of Simplicity Reward during training.
     }
     \label{8}
 \end{figure}

\begin{figure}[!t]
    \centering
     \includegraphics[width=0.9\linewidth]{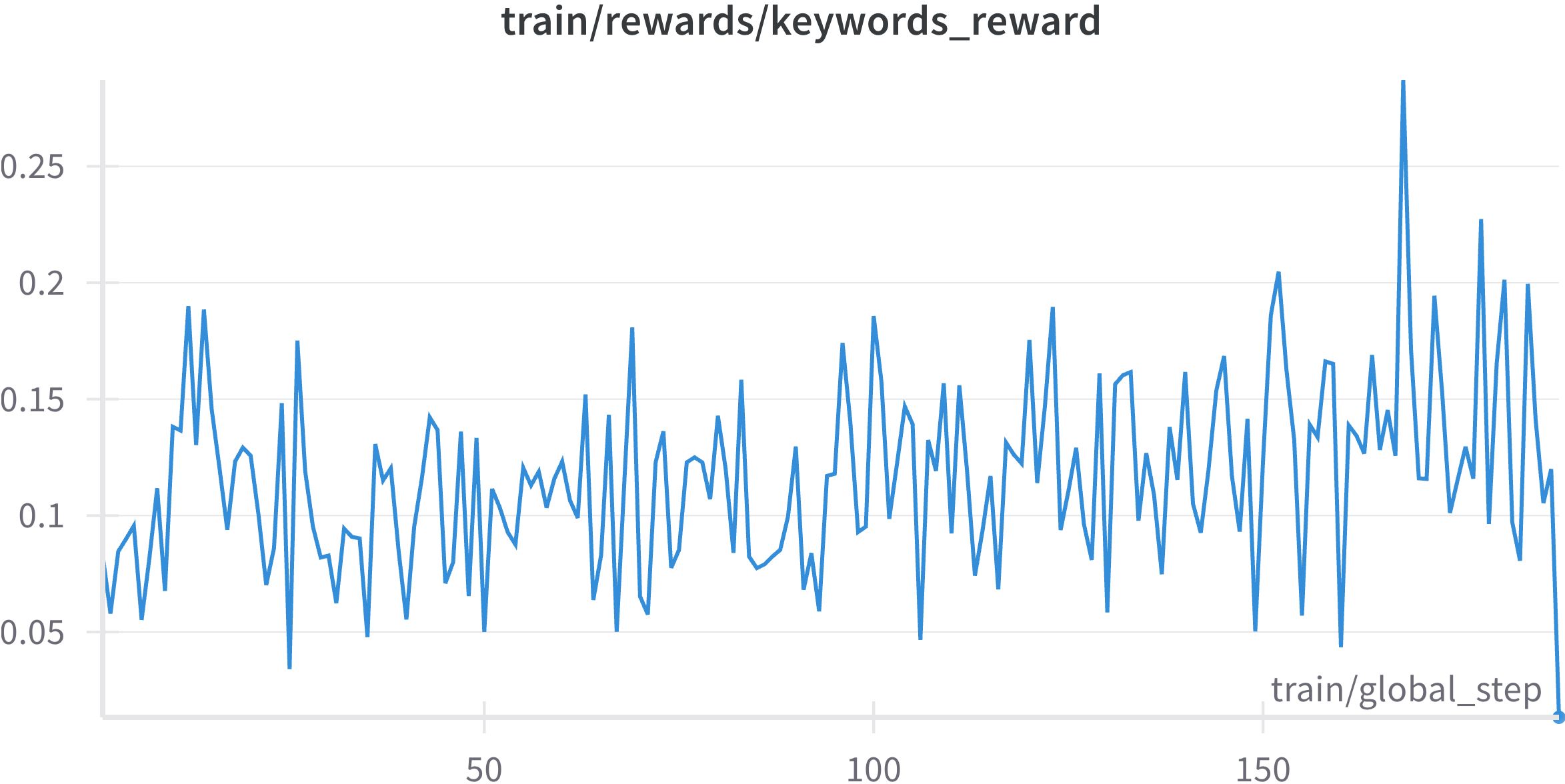}
     \caption{
The variation of Keywords Reward during training.
     }
     \label{9}
 \end{figure}

Figures \ref{5},\ref{6},\ref{7}, \ref{8} and \ref{9} show the changes in the overall reward, accuracy reward, fluency reward, simplicity reward, and keywords reward during the training process. From these plots, it is evident that:

The overall reward curve (Figure \ref{4}) demonstrates a steady increase throughout the training process, reflecting the model’s continuous improvement in generating optimal outputs. The gradual upward trend suggests that the model is effectively learning to maximize its total reward, with occasional fluctuations indicating refinement phases where the model fine-tunes its responses. Similarly, the accuracy reward (Figure \ref{5}) also shows a consistent rise, with some noticeable spikes, especially in the earlier stages of training. These spikes indicate moments of significant improvement in accuracy, likely due to key adjustments in the model's parameters that allow it to better align with the target outputs. The overall trend, however, remains upward, suggesting that the model is steadily improving in generating more accurate results. In Figure \ref{6}, the fluency reward follows a similar trajectory, with a steady climb as training progresses. The fluency reward highlights the model’s ability to produce more coherent and natural outputs, and the stable increase signifies that the model is getting better at generating fluent and contextually appropriate responses over time. The simplicity reward (Figure \ref{7}) shows a sharp initial increase, followed by a period of stabilization and slight oscillations. This suggests that the model quickly learns to generate simpler, more straightforward outputs, but as training progresses, the reward stabilizes, indicating that the model has found a balance between simplicity and complexity in its responses. Finally, the keywords reward (Figure \ref{8}) exhibits noticeable fluctuations but also a general upward trend. This reflects the model's improvement in incorporating relevant keywords into its responses. The sharp spikes in the curve may indicate specific moments where the model made significant improvements in keyword relevance, further enhancing its overall performance.

Overall, these reward curves collectively highlight that the model is making steady progress in optimizing for various aspects, including accuracy, fluency, simplicity, and keyword relevance, demonstrating a well-rounded improvement across different performance metrics throughout the training process.

\subsection{C2: Training Details of the EAD Module}
The EAD module is integrated between the vision encoder and the large language model (LLM) of the VLM. It leverages the visual features extracted by the vision encoder to assess the danger level of the current scene and determines whether a reminder should be triggered. To train the EAD module, we selected 20 real-world walking videos, each representing different outdoor environments. These videos were sampled at regular intervals, and each frame was manually labeled with a danger level, categorized into three levels: A (low), B (medium), and C (high). 
For the training process, we froze the vision encoder from the WalkVLM-LR model and connected it to the EAD module. The training was performed using a fully supervised fine-tuning approach, where the model was trained for four epochs. We used a combination of cross-entropy loss and focal loss \cite{focal_loss} as the supervision signals to ensure that the model learns to differentiate between the varying levels of scene danger effectively. The use of focal loss helps to address class imbalance, especially since lower-risk frames (label A) are more prevalent in the dataset compared to higher-risk frames (label B and C). The EAD module was specifically trained to classify frames based on their danger level, and this process enables it to dynamically evaluate the visual context in real-time. By fine-tuning this module, the VLM can better understand and assess safety-critical situations, which is essential for its real-time application in assisting BLV users during outdoor time.

\section{Appendix D: Challenges and Future Directions}
\subsection{D1: Failure sample analysis}
\begin{figure}[!t]
    \centering
     \includegraphics[width=0.9\linewidth]{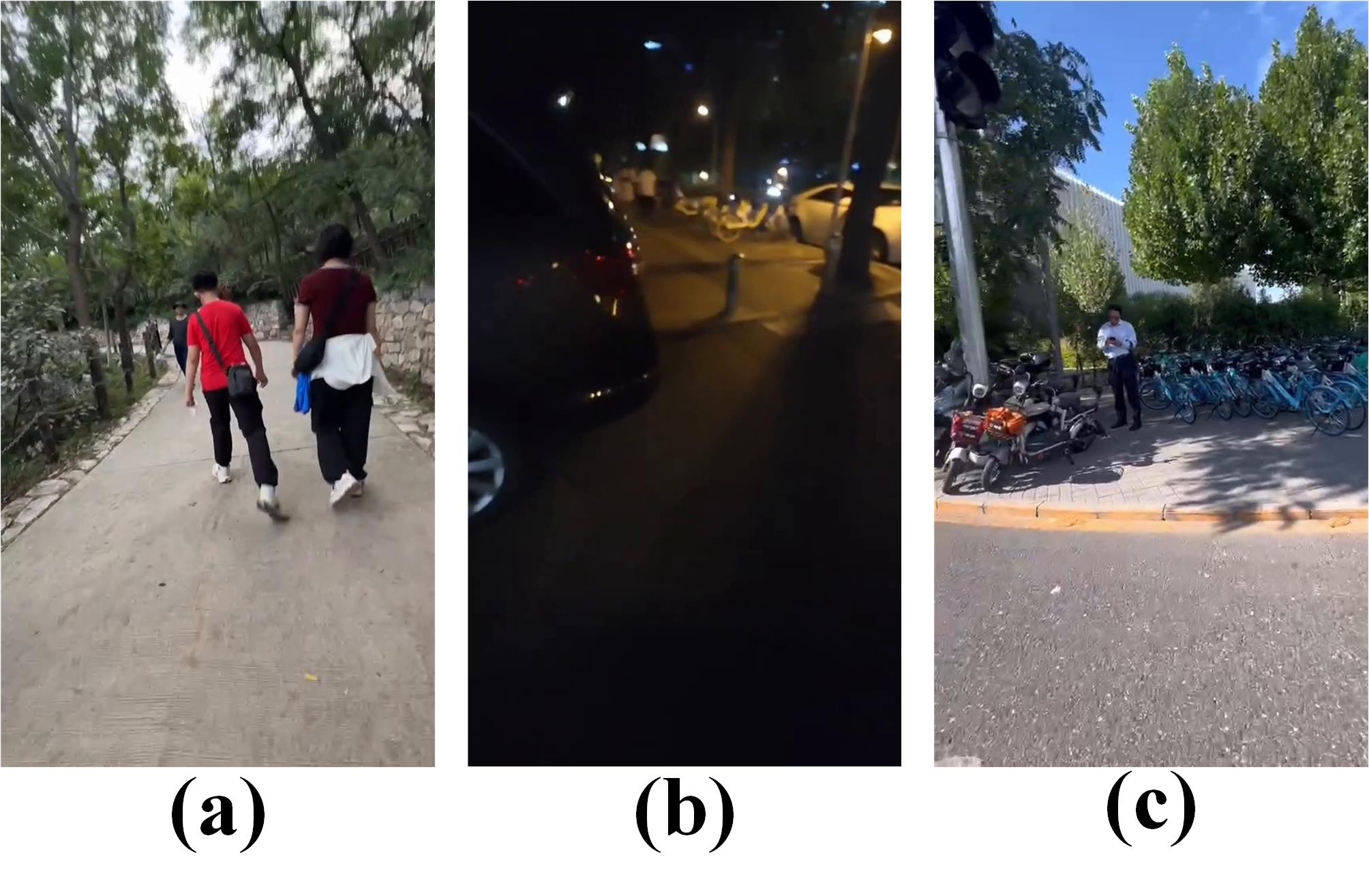}
     \caption{
Examples of WalkVLM Underperformance. These scenarios include streets with heavy pedestrian traffic, roads with low brightness at night, and complex environments with cluttered targets.
     }
     \label{10}
 \end{figure}

To better understand the limitations of WalkVLM-LR, we analyzed its failure cases during the reminder generation process and summarized several recurring issues. Representative examples are shown in Figures \ref{10}.

\subsubsection{Incomplete target coverage and directional misjudgment.}
As illustrated in Figure \ref{10} (a), multiple pedestrians are present at the user’s eleven o’clock, twelve o’clock, and one o’clock directions, all requiring avoidance. However, the generated reminder is:
\textbf{“There is a pedestrian at two o’clock direction, be cautious.”}
This output reveals two issues. First, the model fails to capture all relevant targets, resulting in incomplete situational awareness. Second, the predicted direction of the detected pedestrian is incorrect, potentially misleading users about the true location of obstacles. Such errors are critical for walking assistant safety.

\subsubsection{Inadequate focus on proximate obstacles.}
Figure \ref{10} (b) presents a failure case in a low-light environment. The generated reminder is:
\textbf{“There is a car on the left side of the road, and a bicycle on the right side. Continue straight to avoid any obstacles.”}
While WalkVLM-LR successfully identifies the distant vehicle on the left, it fails to detect and report nearby obstacles within the immediate walking path. This indicates a limitation in spatial attention prioritization, where the model overemphasizes distant and more visually salient objects while neglecting proximate, safety-critical hazards. In real-world applications, this behavior could reduce the practical utility of the reminders.

\subsubsection{Verbose but uninformative reminders in complex scenes.}
As shown in Figure \ref{10} (c), when the visual scene becomes more complex, the generated reminders often become excessively verbose. For example:
\textbf{“There is a group of bicycles parked in the background, with one in particular being closer to the foreground. The bicycles are at a distance, so it's important to maintain a safe distance from them.”}
Although the output is lengthy, it fails to highlight the most critical hazard—an approaching pedestrian directly ahead. This reflects a tendency of the model to produce descriptive but low-priority information in cluttered scenes, while overlooking threats that are immediately relevant to safe walking assistance.

In summary, the failure cases of WalkVLM-LR can be attributed to three main issues: incomplete detection of safety-critical objects, inaccurate spatial localization, and a lack of priority reasoning in complex environments. These observations suggest that future improvements should focus on enhancing close-range obstacle detection, improving spatial grounding, and integrating an attention mechanism that prioritizes imminent threats over background details.

\subsection{D2: Deployment and Challenges}
To enhance accessibility, privacy, and reliability for blind and low-vision (BLV) users, we aim to deploy WalkVLM-LR on edge devices such as smartphones, AR glasses, and wearable assistants. Edge-side deployment eliminates reliance on network connectivity, reduces inference latency, and ensures local processing of sensitive visual data. However, running a 2B-parameter vision-language model (VLM) on-device remains technically challenging, with key considerations summarized below.

\subsubsection{Model Size and Compression Constraints.}
WalkVLM-LR is built upon Qwen2-VL-2B-Instruct, a 2B-parameter VLM. In full-precision formats, the model occupies several gigabytes of storage, exceeding the capacity of many mobile devices. Even with 4-bit quantization (e.g., GPTQ-Int4 or AWQ), the model still consumes around 2.9 GB of GPU memory, posing challenges for integration and thermal management. Efficient quantization, mixed-precision kernels, and selective layer offloading are crucial for practical deployment on resource-constrained platforms\cite{transformer-lite-2024}. 

\subsubsection{Inference Speed and Real‑Time Responsiveness.}
Real-time responsiveness is essential for walking assistance applications. On server-grade GPUs, WalkVLM-LR achieves varying decoding speeds based on the quantization method used. With BF16, the model achieves 35.29 tokens per second, while GPTQ-Int4 achieves the highest speed of 39.76 tokens per second. GPTQ-Int8 and AWQ result in slightly lower speeds of 28.59 and 29.89 tokens per second, respectively (see Table 1). For edge devices, performance varies significantly. On a MacBook Pro M4, decoding speed reaches 71.73 tokens per second with AWQ, making it suitable for real-time processing. However, on a Mac Mini M4 Pro with AWQ, the decoding speed drops to 5.5 tokens per second, indicating that lower-end devices may struggle with real-time processing for walking assistance applications. Considering typical walking-assistant prompts are under 20 tokens, on-device inference for optimized setups can achieve response times within 1–2 seconds, making it feasible for practical use in real-world scenarios \cite{flash-vl-2b-2025}.

\begin{table}[t]
\caption{ Inference Speed for WalkVLM-LR.}
\centering
\begin{tabular}{ccc} 
\toprule
Device             &Quantization & Decoding Speed                \\ 
\hline
A100   & BF16  & 35.29 tokens/s                 \\ 
A100   & GPTQ‑Int4  & 39.76 tokens/s                 \\
A100   & GPTQ-Int8   & 28.59 tokens/s                 \\
A100   & AWQ  & 29.89 tokens/s                  \\
Mac Mini M4 Pro   & AWQ  & 5.5 tokens/s                  \\
MacBook Pro M4   & AWQ  & 71.73 tokens/s                  \\
\bottomrule
\end{tabular}
\end{table}

\subsubsection{Hardware Constraints and Platform Compatibility.}
Successful edge deployment requires devices with 3–4 GB available RAM, a capable mobile GPU or NPU, and support for low-bit kernel execution \cite{hardware_constraints}. High-end smartphones and AR glasses equipped with chipsets like Snapdragon XR2, Apple M-series, or modern MediaTek processors can potentially host VLM inference. Nonetheless, platform heterogeneity introduces integration challenges across TensorFlow Lite, Core ML, and PyTorch Mobile ecosystems. In some cases, native bridging or hybrid runtimes are required for stable and efficient execution \cite{platform_compatibility}.

\subsubsection{Scene Complexity and Model Robustness.}
Outdoor walking scenarios are inherently dynamic, involving varying illumination, occlusions, and crowd densities. WalkVLM-LR exhibits the same limitations as other VLMs in fine-grained spatial reasoning, particularly under complex 3D layouts. Improving real-world robustness may require \cite{scene_complexity}:
\begin{itemize}
\item Integrating depth sensing or stereo cameras for proximity awareness.
\item Leveraging SLAM \cite{slam_integration} or visual-inertial odometry to enhance spatial mapping.
\item Expanding training datasets to cover diverse lighting, weather, and urban conditions.
\end{itemize}

\subsubsection{Summary}
Although deploying a 2B-parameter VLM fully on edge devices is challenging, ongoing advances in model quantization, mobile inference runtimes, and adaptive attention mechanisms make real-time, offline walking assistance increasingly feasible. Achieving this goal would provide BLV users with private, low-latency, and context-aware walking assistance support directly on consumer devices.

\end{document}